\documentclass[lettersize,journal]{IEEEtran}
\usepackage{amsmath,amsfonts}
\usepackage{algorithmic}
\usepackage{array}
\usepackage[caption=false,font=normalsize,labelfont=sf,textfont=sf]{subfig}
\usepackage{textcomp}
\usepackage{stfloats}
\usepackage{url}
\usepackage{verbatim}
\usepackage{graphicx}
\usepackage{hyperref}
\usepackage{booktabs}
\usepackage{tabularx}
\usepackage{multirow}
\usepackage{graphicx}
\usepackage[ruled,vlined,linesnumbered]{algorithm2e}
\newcommand{\codestr}[1]{\texttt{\detokenize{#1}}}

\usepackage[table]{xcolor} 

\def\BibTeX{{\rm B\kern-.05em{\sc i\kern-.025em b}\kern-.08em
    T\kern-.1667em\lower.7ex\hbox{E}\kern-.125emX}}
\usepackage{balance}
\begin{document}
\title{LLM-Driven Preference Data Synthesis for Proactive Prediction of the Next User Utterance in Human–Machine Dialogue}
\author{
Jinqiang Wang, 
Huansheng Ning, ~\IEEEmembership{Senior Member,~IEEE,}
Jianguo Ding,  ~\IEEEmembership{Senior Member,~IEEE,}
Tao Zhu,  ~\IEEEmembership{Senior Member,~IEEE,}
Liming Chen,  ~\IEEEmembership{Senior Member,~IEEE,}
and Chris Nugent  ~\IEEEmembership{Member,~IEEE}
\thanks{Jinqiang Wang and Huansheng Ning are with the School of Computer \& Communication Engineering, University of Science and Technology Beijing, 100083 China. e-mail: jqwang@xs.ustb.edu.cn, ninghuansheng@ustb.edu.cn. \emph{(Corresponding authors: Huansheng Ning and Jianguo Ding)}}
\thanks{Jianguo Ding is with the Department of Computer Science, Blekinge institute of Technology, Karlskrona, Sweden. e-mail: jianguo.ding@bth.se.}
\thanks{Tao Zhu is with the School of Computer Science, University of South China, 421001 China. e-mail: tzhu@usc.edu.cn.}
\thanks{Liming Chen is with the School of Computer Science and Technology, Dalian
University of Technology, Dalian 116024, China.}
\thanks{Chris Nugent is with the School of Computing, Ulster University, Belfast, Northern Ireland, UK.}

\thanks{The code and dataset can be found at: \url{https://github.com/jqwangai/ProUtt}}}


\maketitle

\begin{abstract}
Proactively predicting a user’s next utterance in human–machine dialogue can streamline interaction and improve user experience. Existing commercial API–based solutions are subject to privacy concerns while deploying general-purpose LLMs locally remains computationally expensive. As such, training a compact, task-specific LLM provides a practical alternative. Although user simulator methods can predict a user’s next utterance, they mainly imitate their speaking style rather than advancing the dialogue. Preference data synthesis has been investigated to generate data for proactive next utterance prediction and help align LLMs with user preferences. Yet existing methods lack the ability to explicitly model the intent reasoning that leads to the user’s next utterance and to define and synthesize preference and non-preference reasoning processes for predicting the user’s next utterance.
To address these challenges, we propose ProUtt, an LLM-driven preference data synthesis method for proactive next utterance prediction. ProUtt converts dialogue history into an intent tree and explicitly models intent reasoning trajectories by predicting the next plausible path from both exploitation and exploration perspectives. It then constructs preference and non-preference reasoning processes by perturbing or revising intent tree paths at different future turns. Extensive evaluations using LLM-as-a-judge and human judgments demonstrate that ProUtt consistently outperforms existing data synthesis methods, user simulators, and commercial LLM APIs across four benchmark datasets. We release both the code and the synthesized datasets to facilitate future research.
\end{abstract}


\section{Introduction}
\IEEEPARstart{R}{ECENTLY}, Large Language Models (LLMs) have achieved impressive performance in human–machine dialogue systems \cite{10832392} across domains such as education \cite{zhang2025simulating}, healthcare \cite{wang2024survey}, and tourism \cite{wei2025ctourllm}. Moreover, the proactivity of LLMs has become an increasingly prominent research topic and is often viewed as an important indicator of progress toward Artificial General Intelligence (AGI) \cite{lu2025proactiveagent}. Proactivity encompasses a variety of capabilities, including clarifying ambiguous user queries \cite{wang2025data, qian2024tell}, autonomously executing tasks \cite{lu2025proactiveagent}, and predicting the user's next utterance \cite{wang-etal-2025-know}, a task that is commonly instantiated as a ``you may want to ask'' module. This work focuses on the task of proactively predicting the user's next utterance. By presenting users with candidate next utterances, the system enables selection-based interaction rather than manual text entry, thereby improving user experience.
\par
\begin{figure}
    \centering
    \includegraphics[width=1\linewidth]{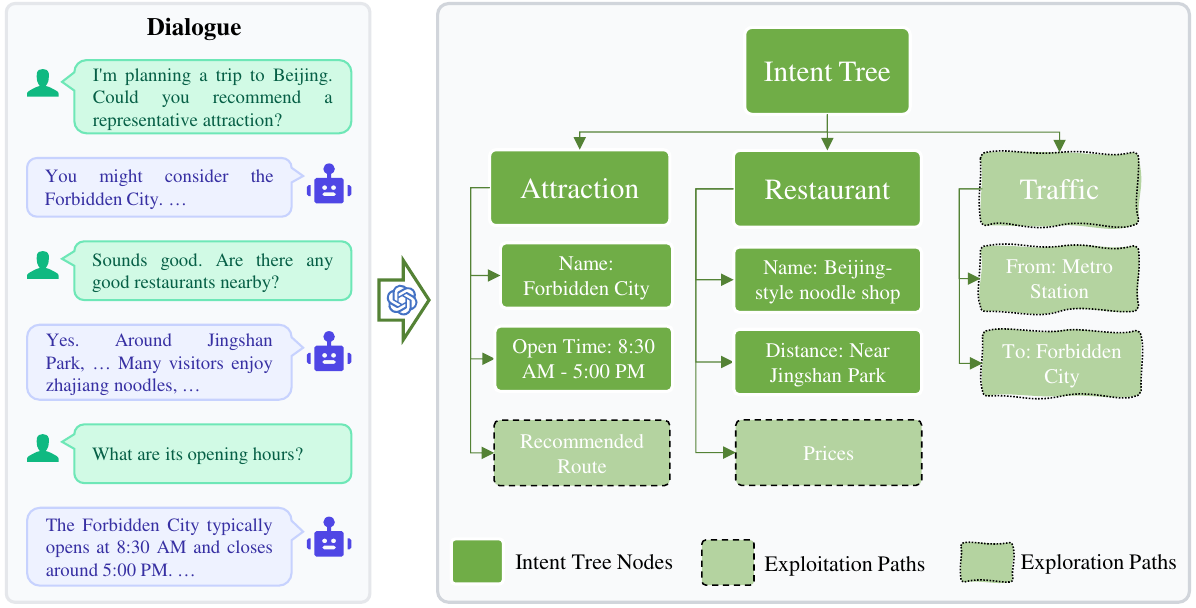}
    \caption{The intent tree extraction process. The LLM converts a human–LLM dialogue into a user intent tree and predicts the next path on the tree from both exploitation and exploration perspectives.}
    \label{fig:tree}
\end{figure}
A straightforward approach to proactive next utterance prediction is to prompt an LLM to analyze the dialogue history and directly predict the user’s upcoming utterance. While this approach can be implemented by calling commercial APIs, cloud-based inference raises significant privacy concerns in sensitive domains such as healthcare and finance, as it requires transmitting raw or intermediate data to third-party platforms \cite{das2025security}. An alternative is to deploy a general-purpose LLM locally to mitigate privacy risks. However, LLMs that demonstrate strong reasoning performance typically contain hundreds of billions of parameters, incurring substantial computational costs that are prohibitive for most institutions \cite{zhu2024survey}. In contrast, lightweight LLMs are more affordable to deploy but often lack reasoning capabilities that are well aligned with the requirements of this task.
\par
Given the above limitations, a more practical direction is to train a task-specific small-scale LLM with fewer than 10B parameters. Existing studies  \cite{wang-etal-2025-know, kong2024platolm} primarily rely on LLM-based user simulators to address this task. These models are typically optimized with respect to the user response loss rather than the assistant objective. Therefore, these approaches tend to overemphasize surface linguistic patterns and user speaking styles, rather than modeling the evolution of user intent that drives the dialogue forward, which can hurt generalization in real-world settings.
Moreover, they lack explicit reasoning mechanisms, preventing them from leveraging the reasoning capabilities of LLMs to predict the user’s next utterance from an intent-centric perspective.
\par
Given the limitations of existing methods, this work adopts a data synthesis paradigm to enable proactive prediction of the user's next utterance. By generating high-quality training data with explicit reasoning traces and leveraging preference learning, data synthesis can both mitigate the shortcomings of existing methods and better align model behavior with human preferences \cite{ouyang2022training}.
To synthesize preference data suited for this task, two key challenges must be addressed. The first is to explicitly model the intent reasoning process that leads to the user’s next utterance. The second is to define and generate both preferred and non-preferred reasoning trajectories for predicting the user’s subsequent utterance. Existing data synthesis methods primarily focus on constructing general-domain datasets and do not sufficiently account for the characteristics of this task or the challenges described above.
\par
To address the issues above, we present ProUtt, an LLM-driven preference data synthesis method for proactively predicting the user’s next utterance in human–LLM interaction. ProUtt innovatively transforms existing multi-turn human–machine dialogue data (e.g., ShareGPT) into a preference dataset enriched with explicit user intent reasoning. The synthesized data guide downstream models to learn intent-level prediction of the next utterance instead of mimicking superficial user style. As shown in Fig.~\ref{fig:tree}, ProUtt first draws on speech act theory \cite{searle1969speech} and goal systems theory \cite{kruglanski2018theory} to prompt the LLM to represent user intentions in multi-turn dialogues as a goal–subgoal tree. Then, inspired by information foraging theory \cite{pirolli1999information}, which posits that people alternate between exploiting and exploring available information sources when planning their next step, ProUtt infers the user’s potential next intent by exploring and exploiting this intent tree, and further converts the inferred intent into a concrete next utterance. This explicit modeling of the user’s intent reasoning trajectory addresses the first challenge of preference data synthesis.
\par
Next, we construct preference samples based on the semantic similarity between the predicted and the gold user utterance. Let $t$ denote the current turn. If the predicted utterance closely matches the gold utterance, we label its corresponding reasoning trajectory as a preferred trajectory. We then perturb this trajectory using intent tree paths from future turns $t+i$ (with $i \ge 2$) to obtain non-preferred trajectories. If the predicted utterance differs substantially from the gold utterance, we directly treat the reasoning trajectory corresponding to the predicted utterance as a non-preferred trajectory. We then revise the predicted reasoning trajectory using the intent tree path from turn $t+1$ and label it as the preferred trajectory.
The above strategy addresses the second challenge. By revising and perturbing trajectories with future intent tree paths, our approach avoids perspective leakage by preventing premature access to the ground-truth answer and guides the downstream model toward intent-level reasoning rather than surface-level phrasing.
\par
We construct preference datasets with explicit reasoning traces for next utterance prediction using ProUtt, based on the English open-domain dialogue dataset LMSYS and the Chinese task-oriented dialogue dataset CrossWOZ. The resulting datasets are named LMSYS-ProUtt-2K, LMSYS-ProUtt-10K, CrossWOZ-ProUtt-2K, and CrossWOZ-ProUtt-5K. The two 2K-sized datasets are used in our experiments to fine-tune a LLM (Qwen3-8B by default), and we use the resulting model performance to validate the effectiveness of our data synthesis method. The remaining larger-scale datasets will be publicly released to facilitate future research. To more comprehensively verify the effectiveness of ProUtt, we further evaluate the fine-tuned model on two real-world dialogue datasets, ShareGPT and WildChat. Importantly, we do not perform any training on these two datasets and use them solely for external generalization testing. For evaluation metrics, we employ LLM-as-a-Judge with both pointwise scoring and pairwise comparison, human evaluation, and semantic similarity analysis based on sentence embeddings. The results show that, compared with existing commercial models (e.g., Qwen3-Max), user simulators, and other data synthesis methods, our proposed synthesis method achieves superior performance across multiple metrics.
\par
Our contributions are summarized as follows:
\begin{enumerate}
\item We propose ProUtt, a novel LLM-based preference data synthesis method that converts multi-turn dialogues into preference reasoning datasets for next utterance prediction.
\item Empirical results indicate that ProUtt-generated preference data enable small-scale LLMs to achieve superior performance on proactive next utterance prediction relative to prior methods across four benchmark datasets, thereby demonstrating the effectiveness of training task-specific small-scale LLMs for this task.
\item To alleviate the scarcity of reasoning-aware preference data for proactive next utterance prediction, we release two preference datasets with explicit reasoning, LMSYS-ProUtt-10K and CrossWOZ-ProUtt-5K, together with the data synthesis code.
\end{enumerate}
\par
The remainder of this paper is organized as follows. Section II reviews related work. Section III presents the proposed preference data synthesis method, ProUtt. Section IV reports and analyzes the main experimental results. Section V provides further discussion on the mechanisms and components of ProUtt. Section VI concludes the paper and outlines directions for future research.

\section{Related Work}
\subsection{Data Synthesis}
Recent evidence suggests that post-training effectiveness depends more on data quality than model scale \cite{lai2025survey}. High-quality synthetic data can reduce privacy and annotation costs while improving performance in dialogue and reasoning settings \cite{long2024llms}. Preference data is central to alignment methods such as RLHF \cite{ouyang2022training} and DPO \cite{rafailov2023dpo}, and its coverage and label fidelity largely determine alignment quality. Accordingly, synthesizing high-quality preference data has become an important direction.
\par
In dialogue scenarios, a common form of LLM data synthesis is multi-turn self-play dialogue generation. Baize \cite{xu2023baize} explores two representative paradigms for this setting, including generating complete multi-turn dialogues in a single pass and constructing dialogues turn by turn by alternating user and assistant roles. UltraChat \cite{ding2023enhancing} adopts a dual-agent setting, where one ChatGPT instance acts as a user to generate queries and the other acts as an assistant to produce responses, enabling scalable construction of interactive dialogue data without human intervention.
For reasoning tasks, recent work has proposed specialized techniques for synthesizing long-CoT data. NuminaMath-CoT \cite{li2024numinamath} samples problems from challenging benchmarks and employs strong reasoning models to generate full long-CoT solutions. In contrast, BOLT \cite{pang2025bolt} introduces a bootstrapping framework that generates long-CoT data without relying on external teacher models, starting from a small set of human-written examples.
\par
Preference data is essential for aligning LLMs, and recent studies have explored using LLMs to automatically construct high-quality preference pairs to reduce manual annotation costs. UltraFeedback \cite{pmlr-v235-cui24f} provides a representative pipeline by combining multi-model sampling with GPT-4-based scoring to obtain fine-grained preference labels. Building on this line, SPaR \cite{cheng2025spar} mitigates spurious differences by refining preference pairs so that suboptimal and optimal responses differ primarily in instruction following. SynPO \cite{dong2025synpo} further forms a closed-loop process that generates tasks, produces candidate responses of varying quality, and performs self-comparison to derive preferences. Magpie \cite{xu2025magpie} extracts large-scale instruction-response pairs from aligned models and constructs preference pairs by scoring and ranking candidate responses using either the same or a stronger model.

\par
These data synthesis methods are designed for general domains and are not well suited for predicting the user’s next utterance. Moreover, existing preference data synthesis approaches do not provide explicit reasoning processes or user intent modeling. Methods such as SPaR and SynPO further require iterative use of intermediate models, which increases synthesis complexity. Our method ProUtt avoids iterative procedures and only requires specifying an LLM configuration. It converts multi-turn dialogue datasets into next-utterance prediction datasets, augmented with explicit preference reasoning processes that model user intent.

\subsection{Proactivity in LLMs}
Most agent systems built on LLMs still rely on explicit human instructions to initiate and execute tasks and typically remain in a passive waiting mode until new instructions arrive \cite{wang2025learning, sahay2025ask}. This paradigm limits the model’s ability to offer assistance or provide autonomous services when no clear request is given. Recent work has begun to explore proactivity in LLMs, with two emerging directions. One line of research focuses on proactively clarifying ambiguous user utterance and the other investigates proactively predicting the user’s subsequent utterance.
\par

For proactive clarification of ambiguous user utterances, prior work treats clarification as an explicit decision-making or planning component in the interaction process. Proactive CoT \cite{deng2023prompting} integrates the decision of whether to provide clarification or guidance into the reasoning chain. Ask Before Plan \cite{zhang2024ask} further formulates clarification as a pre-planning stage and introduces a benchmark together with a multi-agent framework for clarification decision-making and execution. Qian et al. \cite{qian2024tell} propose the IN3 benchmark to provide a systematic evaluation of proactive clarification in LLMs. It decomposes the task into fuzzy intent identification, completion of missing information, and intent summarization, with evaluation cases generated through interactions between user and assistant agents. Building on this direction, SynPT \cite{wang2025data} further enhances proactive clarification by incorporating intent reasoning, affect inference, and candidate option generation to handle ambiguous user utterances more comprehensively.

\par
For proactive prediction of the user’s next utterance, existing research primarily relies on LLM-based user simulators. Kong et al. \cite{kong2024platolm} train a simulator named Socratic on the ShareGPT dataset by treating user turns as training targets while excluding assistant turns, enabling the model to generate the next user utterance conditioned on the dialogue context. Building on this direction, Wang et al. \cite{wang-etal-2025-know} introduce the notion of a user profile and develop a profile generator together with a user simulator called USP, which is further optimized with PPO. The reward is determined by the consistency between the user profile before and after incorporating the predicted next utterance. When the two profiles diverge significantly, the model receives a smaller reward.
\par
However, LLM-based user simulators face overfitting risks and often capture overly fine-grained details in user responses. For example, a model may learn to mimic requests such as “suggest apartments with monthly rent around 1,320–1,380 dollars,” where the specific range carries a long-tail nature and lacks general applicability. Moreover, short dialogues make it difficult to extract stable and reliable user profiles, especially when only a single turn is available. Profile constraints can also suppress the model’s ability to generate diverse user responses and reduce its capacity to introduce new topics. Therefore, our synthesis method does not directly fit highly detailed user behaviors. Instead, it learns more general patterns of user intent trajectories, improving robustness and enabling the model to better cover common user needs.

\section{Method}
\label{sec:method}
\subsection{Overview}
\begin{figure*}
    \centering
    \includegraphics[width=1\linewidth]{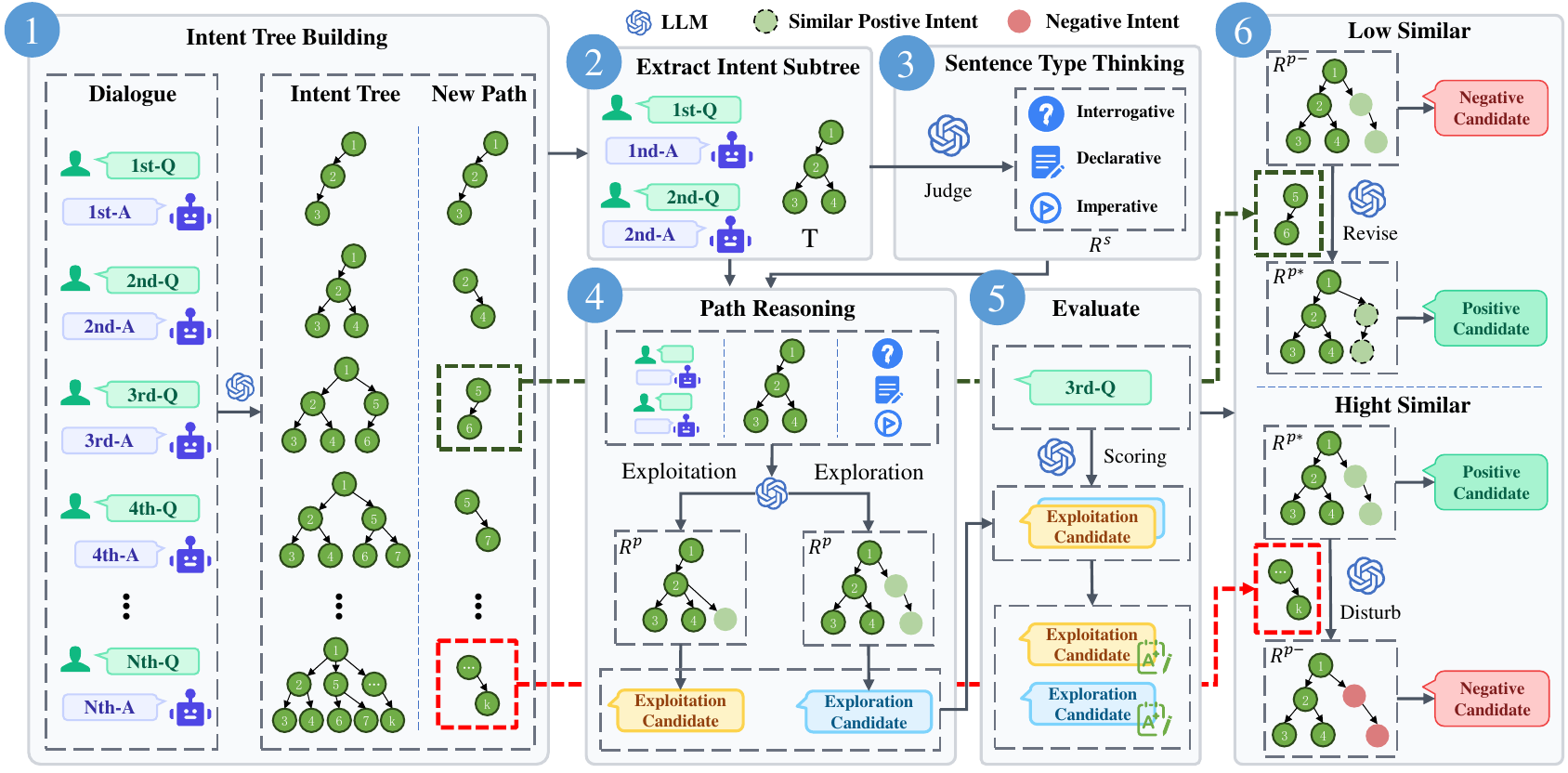}
    \caption{Overview of the ProUtt pipeline, showing hierarchical intent tree construction, sentence type reasoning, intent path reasoning, and the generation of preferred and non-preferred reasoning processes. “Similar Positive Intent” denotes cases where the generated user utterance aligns with the ground-truth intent at a semantic level, without requiring exact surface-form matching.}
    \label{fig:method}
\end{figure*}

Fig. \ref{fig:method} presents the overall workflow of ProUtt, an LLM-driven preference data synthesis method for proactive prediction of the user's next utterance. In the first step, ProUtt prompts the LLM to convert a human–LLM dialogue segment $D$ into a hierarchical intent tree $T$. As shown in the first two components of Fig. \ref{fig:method}, the intent tree provides a structured representation of user intent trajectories across multiple turns. The second step generates the sentence type reasoning $R^s$ for predicting the user’s next utterance, illustrated in the third component of Fig. \ref{fig:method}. ProUtt prompts the LLM to produce candidate reasoning processes for three sentence types: declarative, imperative, and interrogative. We then select the reasoning process whose type best matches the user’s actual next utterance as the preferred sentence type reasoning $R^{s*}$. In addition, we sample one candidate whose type does not match the ground truth as the non-preferred sentence type reasoning $R^{s-}$. The third step constructs the intent path reasoning $R^p$ for predicting the next path of the intent tree, as illustrated in the fourth through sixth components of Fig. \ref{fig:method}. ProUtt prompts the LLM to reason over the intent tree from both exploitation and exploration perspectives and to propose the next plausible path $R^p$, which is then used to generate a predicted next user utterance. When the predicted utterance closely matches the ground truth, the corresponding reasoning process is treated as the preferred intent path reasoning $R^{p*}$. The non-preferred intent path reasoning $R^{p-}$ is obtained by perturbing $R^{p}$ using intent tree paths from future turns beyond the next one. If the predicted utterance diverges substantially from the ground truth, the reasoning process is labeled as the non-preferred intent path reasoning $R^{p-}$. In this case, the preferred intent path reasoning $R^{p*}$ is produced by revising $R^{p}$ using the immediate next path of the intent tree.
\par
Finally, ProUtt aggregates these components into complete preference data. Representative examples of the synthesized instances are shown in Appendix Fig.~\ref{fig:appendix-pref-example-chosen} and Fig.~\ref{fig:appendix-pref-example-rejected}. Each instance contains the dialogue context $D$ and its associated intent tree $T$, together with a preferred reasoning trajectory $r^+=(R^{s*},R^{p*})$ and a non-preferred reasoning trajectory $r^-=(R^{s-},R^{p-})$. Each trajectory produces a next user utterance by verbalizing the predicted intent path, denoted $\hat{U}^{+}$ and $\hat{U}^{-}$. 
This structured construction allows ProUtt to integrate naturally with preference-based alignment methods. The preferred pair \((r^{+}, \hat{U}^{+})\) and the non-preferred pair \((r^{-}, \hat{U}^{-})\) can be used directly by objectives such as DPO, which encourage downstream models to assign higher likelihood to the preferred reasoning and output. Conceptually, the alignment process aims to enforce
\begin{equation}
\pi_\theta \big(r^{+}, \hat{U}^{+} \mid D, T\big)
\gg
\pi_\theta \big(r^{-}, \hat{U}^{-} \mid D, T\big)
\end{equation}
so that downstream models learn to generate preference-aligned reasoning trajectories and next user utterances.
\par
For clarity, we present a concise algorithmic summary of the overall ProUtt data synthesis pipeline in Algorithm~\ref{alg: algorithm} in the appendix.

\subsection{Intent Tree}
We introduce a strategy that converts multi-turn human–LLM dialogues into hierarchical intent trees for explicitly modeling the dependency structure of user intent. The strategy is inspired by speech act theory \cite{searle1969speech} and goal systems theory \cite{kruglanski2018theory}. Speech act theory emphasizes that utterances convey not only surface meanings but also the speaker’s underlying intentions. Goal systems theory further suggests that human intent often exhibits a hierarchical goal–subgoal structure that unfolds through progressive decomposition during task execution. Building on these insights, we incorporate hierarchical tree structures into dialogue modeling, enabling LLMs to explicitly encode and reason over user intent in a tree-structured form.
\par
Let a multi-turn dialogue be denoted as 
$D^{(N)} = \{ d_i \}_{i=1}^{N}$,
where $N$ is the total number of turns and each turn 
$d_i = \langle u_i, a_i \rangle$ 
consists of the user utterance $u_i$ and the assistant response $a_i$.

As shown in the first part of Fig. \ref{fig:method}, ProUtt first prompts the LLM to convert the dialogue $D^{(N)}$ into a user intent tree 
$T^{(N)} = \{ t_i \}_{i=1}^{N}$ 
that unfolds along the dialogue turns. Based on this tree, we further obtain the newly added intent paths at each turn
$P^{(N)} = \{ p_i \}_{i=1}^{N}$.
\par
As illustrated in Fig. \ref{fig:tree}, the first-level child nodes of the intent tree represent high-level topics such as creative writing or code generation. The second-level nodes specify topic-specific attributes, for example writing style and length constraints, or programming languages, dependency libraries and interface requirements. This hierarchical structure enables the LLM to better capture shifts in the user's semantic focus and provides structural priors for subsequent reasoning.

For data construction, we extract the first $K$ rounds of the dialogue $D^{(K)}$ together with their corresponding intent subtree $T^{(K)}$ to build preference samples with explicit reasoning, where $1 \le K \le N - 1$.
Fig. \ref{fig:method} illustrates the process of extracting intent paths and generating the associated preference data from a multi-turn dialogue, illustrated for the case $K = 2$.

\subsection{Sentence Type}
We introduce a sentence type reasoning mechanism that guides the LLM to explicitly infer the sentence type of the user's next utterance, such as declarative, imperative or interrogative.
This design also draws on speech act theory and emphasizes the user’s underlying action intent. In this formulation, declarative forms correspond to conveying information or stating facts, imperative forms reflect requests or commands, and interrogative forms represent questioning and information-seeking behaviors. Accordingly, ProUtt decouples action intent from the current user utterance and uses it to predict the next user utterance.
\par
Based on this theory, we define the set of sentence types as
\begin{equation}
S = \{ s_1, s_2, s_3 \} = 
\{ \text{declarative}, \text{imperative}, \text{interrogative} \}
\label{eq:stype-set}
\end{equation}
\par

For each sentence type $s_j \in S$, we prompt the LLM, conditioned on the dialogue context $D^{(K)}$, to produce a corresponding reasoning chain, denoted as $r_{s_j}$.

\par
Let $s_{K+1}^{*}$ denote the ground-truth sentence type of the next user utterance $u_{K+1}$. From the candidate reasoning chains, we select the chain matching $s_{K+1}^{*}$ and take it as the sentence type reasoning for the chosen response:

\begin{equation}
R_{K+1}^{s*} = r_{\, s_{K+1}^{*}}
\label{eq:chosen-stype}
\end{equation}
\par
For the rejected response, we randomly sample a sentence type from the remaining types and take the corresponding reasoning chain:
\begin{equation}
R_{K+1}^{s-} = r_{\bar{s}}
\qquad 
\bar{s} \sim \text{Unif}\!\left( S \setminus \{ s_{K+1}^{*} \} \right) 
\label{eq:rejected-stype}
\end{equation}

\subsection{Intent Tree Path Reasoning}
\subsubsection{Path Reasoning}
After sentence type reasoning, ProUtt further generates a reasoning process to predict the next path on the user intent tree $T^{(K)}$. Inspired by information foraging theory \cite{pirolli1999information}, which posits that humans dynamically trade off between exploiting current information sources and exploring new ones to maximize information gain, ProUtt extends this idea into a parallel dual-perspective modeling procedure. It prompts the LLM to infer the next path on the user intent tree $T^{(K)}$ from two complementary perspectives, exploitation and exploration, conditioned on the current dialogue context $D^{(K)}$ and the inferred sentence type.
\par
For each perspective $v \in \{M, E\}$, ProUtt generates $Q$ candidate next-step paths on the user intent tree, denoted as
\begin{equation}
\hat{P}_{K+1}^v = \left\{ \hat{p}_{K+1}^{(v,q)} \right\}_{q=1}^{Q}
\end{equation}
where $M$ denotes the exploitation perspective and $E$ denotes the exploration perspective.
\par
The exploitation perspective focuses on enriching attributes under existing topics in the intent tree or proposing alternative values for existing attributes, which yields $\hat{P}_{K+1}^M$. The exploration perspective instead targets topics and attributes that are not yet present in the intent tree, producing exploratory paths $\hat{P}_{K+1}^E$. Combining the two perspectives results in a total of $2Q$ candidate paths:
\begin{equation}
\hat{P}_{K+1} = \hat{P}_{K+1}^M \cup \hat{P}_{K+1}^E
\end{equation}
\par
We refer to these candidate intent tree paths together with their associated reasoning steps as the intent tree path reasoning process $R_{K+1}^{p}$. Finally, ProUtt prompts the LLM to verbalize each predicted path in $\hat{P}_{K+1}$ into a concrete user utterance given the context $D^{(K)}$, producing
\begin{equation}
\hat{U}_{K+1} = \left\{ \hat{u}_{K+1}^i \right\}_{i=1}^{2Q}
\end{equation}
\subsubsection{Evaluation}
We adopt an LLM-as-a-judge approach to compute the similarity between the ground-truth next utterance $u_{K+1}$ and each predicted utterance in $\hat{U}_{K+1}$, and we take the maximum similarity score $J_{\max} \in [0,1]$. We then define a high-confidence threshold $\tau_{\text{high}}$ and a low-confidence threshold $\tau_{\text{low}}$, and use them to construct preference and non-preference labels for $R_{K+1}^{p}$.
\par
When $J_{\max} \ge \tau_{\text{high}}$, we regard the current intent tree path reasoning process $R_{K+1}^p$ as a preferred reasoning process and denote it by $R_{K+1}^{p*}$, while the non-preferred reasoning process $R_{K+1}^{p-}$ is obtained by perturbation. When $J_{\max} \le \tau_{\text{low}}$, we instead treat the current reasoning process $R_{K+1}^p$ as a non-preferred reasoning process $R_{K+1}^{p-}$ and obtain the preferred reasoning process $R_{K+1}^{p*}$ by revision. When $\tau_{\text{low}} < J_{\max} < \tau_{\text{high}}$, we consider this case uncertain, where $R_{K+1}^{p*}$ is obtained via revision and $R_{K+1}^{p-}$ is obtained via perturbation.
\par
If a reasoning process is labeled as a preferred sample, its predicted user utterance is marked as the preferred output $\hat{U}_{K+1}^*$. If it is labeled as a non-preferred sample, the corresponding predicted user utterance is marked as the non-preferred output $\hat{U}_{K+1}^-$.

\subsubsection{Revision and Perturbation}
Revision and perturbation are applied only to intent tree path reasoning processes that correspond to the same perspective (exploitation or exploration) as the ground-truth next utterance $u_{K+1}$.
\par
For revision, ProUtt uses the ground-truth next intent tree path $p_{K+1}$ and replaces the predicted path in the original intent tree path reasoning process $R_{K+1}^p$, while keeping the remaining reasoning steps unchanged. Importantly, this revision does not directly condition on the ground-truth user utterance $u_{K+1}$. This design avoids leaking information from the ground-truth next utterance $u_{K+1}$ and helps maintain the generalization ability of the synthesis process. Based on the revised reasoning process $R_{K+1}^{p*}$, ProUtt then prompts the LLM to generate an approximately correct next user utterance $\hat{U}_{K+1}^*$. The term “approximately” is intended to weaken overfitting to user-specific writing style and to encourage the model to learn intent-level prediction rather than surface-level phrasing. Therefore, in Fig. \ref{fig:method}, the legend “Similar Positive Intent” reflects this notion of approximate correctness at the intent level rather than exact surface-form matching.
\par
For perturbation, ProUtt uses a more distant future intent tree path $p_{K+\varepsilon}$, where $2 \le \varepsilon \le N-K$, to replace the predicted path in $R_{K+1}^p$, which yields an incorrect next user utterance $\hat{U}_{K+1}^-$. If $p_{K+\varepsilon}$ is detected to be semantically identical or highly similar to $p_{K+1}$, ProUtt instead prompts the LLM, conditioned on $p_{K+1}$, to generate an alternative path that is logically plausible but semantically distinguishable, and uses this LLM-generated path as the perturbed incorrect path.

\section{Experimental Results}

\subsection{Pointwise Evaluation}

Following the experimental setup in Appendix~\ref{sec:exp-setup}, we first conduct a pointwise evaluation of the proposed preference data synthesis method. We report two types of metrics: \textit{LLM-Judge}, an LLM-based evaluator score that measures intent consistency and semantic relevance, and \textit{Embed-Sim}, an embedding-based semantic similarity score. The pointwise results on the four test sets are summarized in Table~\ref{tab:pointwise}.

\begin{table*}[t]
\centering
\small
\caption{Pointwise evaluation results. All values are reported in \%. For our method, we report mean $\pm$ 95\% confidence intervals.}
\label{tab:pointwise}
\resizebox{\linewidth}{!}{
\begin{tabular}{lcccccccc}
\toprule
\multirow{2}{*}{Method} &
\multicolumn{2}{c}{LMSYS} &
\multicolumn{2}{c}{ShareGPT} &
\multicolumn{2}{c}{WildChat} &
\multicolumn{2}{c}{CrossWOZ} \\
\cmidrule(lr){2-3}
\cmidrule(lr){4-5}
\cmidrule(lr){6-7}
\cmidrule(lr){8-9}
& LLM-Judge & Embed-Sim & LLM-Judge & Embed-Sim & LLM-Judge & Embed-Sim & LLM-Judge & Embed-Sim \\
\midrule
\multicolumn{9}{l}{\textbf{Prompt-based LLMs}} \\
Qwen3-8B \cite{yang2025qwen3}             & 45.70 & 69.90 & 46.40 & 67.06 & 38.50 & 64.10 & 37.70 & 53.91 \\
Doubao-1.5-Pro \cite{doubao15pro}        & 53.10 & 72.45 & 46.00 & 68.76 & 49.60 & 71.18 & 40.01 & 51.94 \\
Doubao-Seed-1.6  \cite{doubaoseed16}       & 50.50 & 72.67 & 42.70 & 68.78 & 44.30 & 69.22 & 37.10 & 53.06 \\
DeepSeek-V3.2-Exp \cite{deepseekv32}    & 53.50 & 74.09 & 46.40 & 68.24 & 48.90 & 71.06 & 43.60 & 54.05 \\
GLM-4.6 \cite{glm2024chatglm}               & 56.10 & 72.08 & 49.50 & 69.19 & 47.60 & 70.24 & 37.80 & 53.84 \\
Qwen3-Max \cite{yang2025qwen3}           & 53.10 & 74.32 & 43.60 & 68.85 & 40.80 & 69.57 & 39.00 & 54.04 \\
\midrule
\multicolumn{9}{l}{\textbf{User Simulators}} \\
Socratic \cite{kong2024platolm}         & 49.50 & 72.51 & 32.40 & 66.31 & 39.40 & 69.71 & 40.90 & 53.92 \\
USP \cite{wang-etal-2025-know}           & 38.90 & 70.97 & 27.70 & 66.31 & 31.10 & 67.06 & 14.70 & 48.21 \\
\midrule
\addlinespace[0.3em]
\multicolumn{9}{l}{\textbf{Data Synthesis}} \\
Baize \cite{xu2023baize}            & 55.09 & 74.42 & 46.30 & 68.19 & 48.50 & 73.15 & 46.10 & 53.91 \\
SPaR (SFT) \cite{cheng2025spar}            & 51.20 & 70.86 & 43.90 & 68.93 & 44.90 & 70.20 & 40.90 & 52.91 \\
SPaR (DPO)            & 52.90 & 72.75 & 45.10 & 68.97 & 45.40 & 69.68 & 41.40 & 53.22 \\
MAGPIE (SFT)  \cite{xu2025magpie}        & 54.79 & 71.91 & 44.60 & 68.62 & 47.90 & 70.51 & 38.00 & 53.97 \\
MAGPIE (DPO)          & 55.19 & 71.16 & 44.30 & 68.61 & 49.70 & 70.81 & 39.00 & 53.17 \\
\midrule
\addlinespace[0.3em]
\multicolumn{9}{l}{\textbf{ProUtt (ours)}} \\
SFT
& 58.10$\pm$2.0
& 76.30$\pm$0.5
& 50.12$\pm$1.7
& 69.08$\pm$0.3
& 50.80$\pm$4.5
& 73.22$\pm$0.6
& 45.98$\pm$0.5
& 53.59$\pm$1.0 \\

SFT + DPO
& \textbf{60.98}$\pm$2.4
& \textbf{76.98}$\pm$0.4
& \textbf{52.66}$\pm$2.6
& 69.35$\pm$0.2
& \textbf{52.16}$\pm$1.3
& 73.80$\pm$0.6
& \textbf{51.22}$\pm$1.7
& 53.50$\pm$0.8 \\

SFT + ORPO
& 59.16$\pm$1.2
& 74.17$\pm$3.7
& 48.00$\pm$2.2
& \textbf{69.81}$\pm$0.6
& 50.90$\pm$2.3
& 73.78$\pm$0.7
& 48.28$\pm$0.9
& \textbf{54.15}$\pm$0.4 \\

SFT + SimPO
& 60.56$\pm$1.9
& 73.09$\pm$1.1
& 48.60$\pm$1.8
& 68.53$\pm$0.3
& 50.50$\pm$1.5
& \textbf{73.89}$\pm$0.8
& 46.70$\pm$3.2
& 52.77$\pm$2.2 \\

\bottomrule
\end{tabular}
}
\end{table*}

From the perspective of LLM-based pointwise evaluation, ProUtt achieves the best performance on all four datasets. Specifically, compared with the backbone model Qwen3-8B, ProUtt improves the downstream model performance on LMSYS, ShareGPT, WildChat and CrossWOZ by 15.28\%, 6.26\%, 13.66\% and 13.52\% respectively. This shows that the preference datasets synthesized by ProUtt effectively enhance model performance on proactive next utterance prediction. In addition, compared with the strongest baseline method, ProUtt yields relative gains of 4.88\%, 3.16\%, 2.46\% and 5.12\% on the four datasets, demonstrating the effectiveness of reasoning about the next user utterance via user intent trees. We additionally report 95\% confidence intervals for ProUtt to reflect the statistical reliability of the observed improvements.
\par
Among the prompt-based methods, the open-source models DeepSeek-V3.2-Exp and GLM4.6 achieve strong results, with 685B and 355B parameters, respectively. However, our 8B model fine-tuned with ProUtt surpasses both, indicating that fine-tuning a compact model on ProUtt-synthesized preference data can outperform much larger models for proactive next utterance prediction.
\par
In ProUtt, we use Doubao-1.5-Pro as the default LLM for data synthesis, which can be viewed as an approximate form of distillation from Doubao-1.5-Pro. Nonetheless, the model fine-tuned on ProUtt data outperforms Doubao-1.5-Pro itself, further highlighting the importance of the exploitation and exploration paradigm centered on user intent trees.
\par
The two user-simulator baselines perform poorly. Our inspection of their outputs suggests that they tend to overfit the user’s speaking style, placing insufficient emphasis to reasoning about the user’s next intent.
\par
For the other data synthesis methods, SPaR is relatively weak and relies on a revision strategy conditioned on the ground-truth response. Baize and MAGPIE perform better than SPaR. Baize directly generates data conditioned on the ground truth, whereas MAGPIE mainly adopts a self-consistency-based strategy. We observe that the synthesized data of SPaR and Baize suffer from severe ground-truth perspective leakage, which leads the trained models to produce misleading ``ground-truth-aware’’ descriptions during reasoning. Although MAGPIE effectively avoids such perspective leakage, the absence of a revision mechanism prevents it from fully fitting users’ intent trajectories over large numbers of samples.
\par
In the SFT and alignment experiments, DPO yields consistently stronger performance than SFT, confirming the usefulness of preference data for next-utterance prediction. ORPO and SimPO deliver modest improvements on LMSYS and CrossWOZ, perform on par with SFT on WildChat, and slightly degrade performance on ShareGPT. This indicates that while ProUtt’s preference data can be used with all three alignment algorithms, in practice we recommend DPO as the primary alignment strategy.
\par

From the perspective of embedding-based pointwise evaluation, ProUtt achieves the best results on all four datasets. Compared with Qwen3-8B, ProUtt improves embedding-based similarity scores on LMSYS, ShareGPT, WildChat, and CrossWOZ by 7.08\%, 2.75\%, 9.79\%, and 0.24\%, respectively. The gain on CrossWOZ is relatively small, as indicated by the reported 95\% confidence interval. To better understand this result, we examined CrossWOZ outputs together with the embedding similarity computation. Our analysis suggests that ProUtt emphasizes core user intent and semantic consistency, whereas embedding similarity is computed at the whole-string level and assigns weight to discourse markers, pronouns, and other surface tokens. For example, models trained with ProUtt tend to produce full location names or more explicit expressions, while other models often rely on pronouns or ellipsis. Although this may slightly reduce embedding similarity, it leads to clearer responses in practice and helps reduce user ambiguity.

\subsection{Pairwise Evaluation}
\label{sec:pairwise_evaluation}
We conducted pairwise comparisons between ProUtt (DPO) and four representative baselines. The baselines include two prompt-based methods (Qwen3-8B and DeepSeek-V3.2-Exp), one user simulator (Socratic), and one data synthesis method (MAGPIE). Qwen3-8B serves as the unfine-tuned backbone model that reflects the raw performance, while the other baselines are chosen because they perform strongly under pointwise evaluation.
\begin{figure*}
    \centering
    \includegraphics[width=1\linewidth]{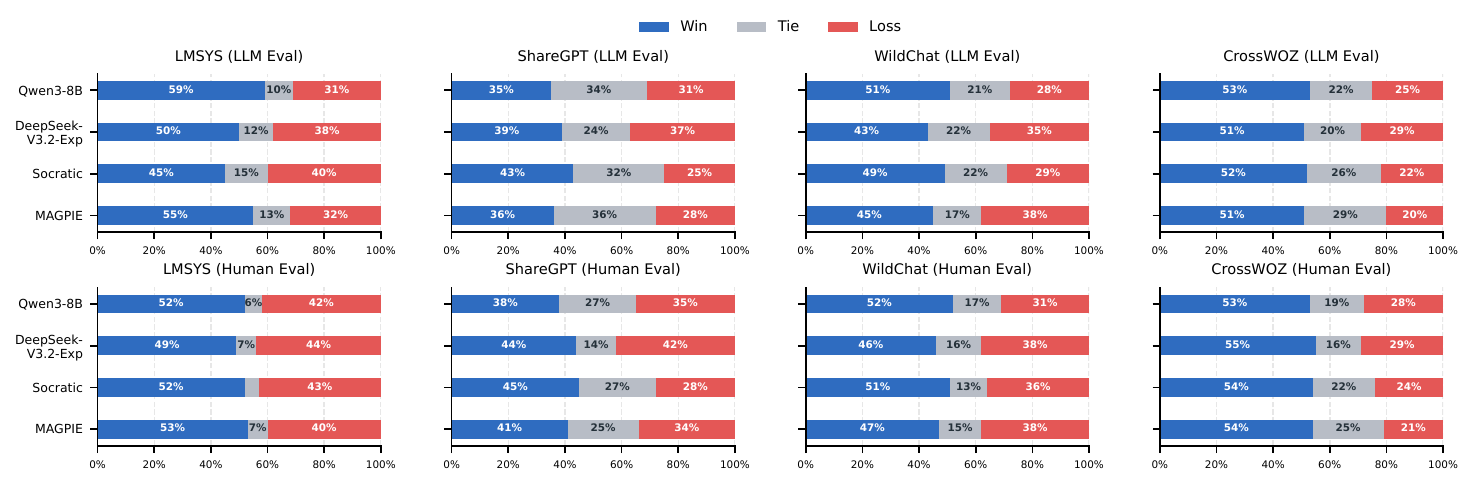}
    \caption{Pairwise win–tie–loss comparison of ProUtt and baselines on four dialogue datasets under LLM and human evaluation.}
    \label{fig:pairwise}
\end{figure*}
\par
\par
On the four datasets LMSYS, ShareGPT, WildChat and CrossWOZ, Fig. \ref{fig:pairwise} reports the win–tie–loss ratios between ProUtt and each baseline under both LLM-based and human evaluations. The results show that ProUtt consistently outperforms all baselines. On CrossWOZ, ProUtt achieves a win rate above 50\%. On LMSYS and WildChat, ProUtt also exhibits a clear advantage over the other methods. On ShareGPT, the improvement is less pronounced. According to feedback from human annotators, test cases in ShareGPT are more complex than those in the other datasets, which degrades the performance of all models and leads to a larger number of ties.
\par
To further assess the reliability of our conclusions, we analyze the consistency between LLM and human evaluations using two metrics, Agreement Rate and Cohen’s $\kappa$. The detailed consistency patterns are visualized with bar plots and confusion matrices in Appendix Fig. \ref{fig:appendix-consistency} and \ref{fig:appendix-confusion}. The figures also report 95\% confidence intervals for Agreement Rate and Cohen’s $\kappa$ in order to illustrate the statistical robustness of the evaluation. The results show that the Agreement Rate between LLM and human judgments exceeds 80\% in all settings. Cohen’s $\kappa$ is generally above 0.6, indicating substantial agreement, and even exceeds 0.8 on CrossWOZ, which reflects a high level of consistency between LLM-based and human evaluations. Consistently, the confusion matrices show that most comparisons lie on the diagonal with relatively few off-diagonal discrepancies, further corroborating the strong agreement between human and LLM judgments. These findings demonstrate that the pairwise evaluation protocol is stable.

\section{Discussions}
\subsection{Ablation Studies}
This subsection conducted ablation studies to quantify the contribution of each key component in ProUtt. Specifically, we examine three modules: the user intent tree, the sentence type reasoning mechanism, and the negative example construction strategy based on future intent tree paths. In the ablation without the intent tree module, the data synthesis pipeline no longer first models user intent as a tree structure and instead directly performs subsequent synthesis based on the dialogue context text. In the ablation without the sentence type reasoning module, ProUtt no longer infers the sentence type of the next user utterance, such as interrogative, declarative, or imperative, and directly proceeds to intent tree path reasoning. In the variant that replaces the negative example construction strategy based on future intent tree paths, we instead prompt the LLM to automatically generate negative intent paths, which serve as non-preferred intent reasoning trajectories. Appendix Table \ref{tab:ablation} presents the experimental results.

\par
From the ablation results, we find that removing the intent tree leads to the most substantial performance degradation on open-domain dialogue datasets, while the impact is smaller on task-oriented datasets. We posit that user intent in open-domain conversations is often implicit. Without an explicit intent modeling process, the LLM becomes less capable of inferring or exploring latent intent, which ultimately lowers overall performance. Removing the sentence type reasoning module also yields consistent performance drops across all datasets. This pattern indicates that anticipating the user’s action-level intent is important for enabling the downstream model to better capture semantic intent. Finally, eliminating the strategy that constructs negative examples based on future intent tree paths reduces performance on every dataset. Although the negative paths generated automatically by the LLM are syntactically plausible, these paths show weak semantic alignment with the dialogue context and often correspond to user trajectories with very low likelihood.

\subsection{Effect of Different LLMs}
\label{sec:llm-backbone}
In this subsection, we investigate how replacing the default LLM used in ProUtt affects the quality of synthesized data. Specifically, we substitute the default model Doubao-1.5-Pro with three alternative LLMs, Qwen3-Plus, DeepSeek-3.2-Exp, and GLM-4.5-Air. The results are summarized in Appendix Table \ref{tab:llm-backbone}.

\par 
The experimental results show that replacing the underlying LLM leads to only minor fluctuations in ProUtt’s overall performance. Across all datasets and metrics, the performance range remains within approximately 2.7 percentage points, indicating that ProUtt exhibits strong robustness and adaptability to different LLM backbones. For open-domain dialogue datasets, performance exhibits slightly higher variance, as these tasks are more sensitive to stylistic variation in LLM-generated responses. In contrast, task-oriented dialogue datasets display smaller variance, as their structured nature allows key elements such as user intents and slot information to be consistently captured across different LLMs. While Doubao-1.5-Pro remains the best-performing backbone on most metrics, the other LLMs surpass it in several cases. This indicates that they offer complementary strengths in terms of linguistic diversity and style, and points to hybrid-LLM data synthesis as a promising direction for future work.
\subsection{Sensitivity of Backbone Models to Synthetic Data}
\label{sec:backbone}
This subsection investigates how backbone LLMs with different parameter scales and from different model families respond to ProUtt-generated synthetic data. Appendix Table \ref{tab:backbone-sensitivity} summarizes the results across four evaluation benchmarks.

\par
We find that the benefits of synthetic data are not uniform across models. Larger models consistently achieve greater improvements, suggesting that increased capacity enables them to better absorb the intent trajectory patterns encoded in the synthesized data. We also observe differences across model families. Although LLaMA3.1-8B \cite{grattafiori2024llama} performs well on datasets aligned with its training distribution, such as LMSYS (LLM-based) and CrossWOZ (embedding-based), it exhibits weaker generalization to out-of-distribution benchmarks. On datasets like ShareGPT and WildChat, LLaMA3.1-8B clearly underperforms Qwen3-8B. This contrast suggests that different LLM families differ in their ability to transfer synthesized intent reasoning patterns to unseen dialogue domains.

\subsection{Impact of Instruction Structure}
\label{sec:instruction}
In this subsection, we examine how the level of instruction structure affects model performance. We consider two variants of task instructions. Structured instructions explicitly present the user intent tree and outline the intermediate reasoning steps required for predicting the next user utterance. Minimal instructions describe the prediction task only at a high level without revealing any intermediate reasoning or intent tree structure. The results are summarized in Appendix Table~\ref{tab:instruction-structure}.

\par
Models trained with minimal instruction templates show a consistent decline in performance. The decline is relatively small on task-oriented datasets and more substantial on open-domain dialogue datasets. To further investigate this effect, we analyze the alignment between model-generated reasoning trajectories and the ground-truth intent tree paths. Both instruction formats allow the model to generate generally reasonable reasoning steps. However, with minimal instructions, the reasoning trajectories tend to deviate from the topics and attributes already present in the intent tree. These deviations are not arbitrary. Instead, they reflect broader exploratory behavior that spans a wider range of plausible reasoning directions. In summary, compared with structured instructions, minimal instructions encourage the downstream model to follow the main direction of the intent tree while engaging in more diffuse exploration of related topics. This leads to slightly lower alignment with the ground-truth path, although the generated reasoning remains semantically coherent.

\subsection{Threshold Sensitivity Analysis}
\label{sec:threshold}
This subsection investigates how two key confidence thresholds in ProUtt, the high-confidence threshold $\tau_{\text{high}}$ and the low-confidence threshold $\tau_{\text{low}}$, affect model performance. These thresholds jointly determine whether an instance is assigned as a positive or negative example, and whether it enters the revision or perturbation stage. In addition to the default configuration, we examine three alternative threshold settings to assess the sensitivity of ProUtt to different choices of $\tau_{\text{high}}$ and $\tau_{\text{low}}$. The results are summarized in Appendix Table~\ref{tab:threshold-sensitivity}.

When the thresholds are set too closely, for example $\tau_{\text{high}} = 0.7$ and $\tau_{\text{low}} = 0.4$, the boundary between positive and negative instances becomes substantially blurred. Under such conditions, instances that only weakly align with the ground-truth user intent are often misclassified as positives, while instances that still retain partial semantic overlap with the ground truth may be categorized as negatives. This weak discrimination hampers supervised learning from accurately fitting intent trajectories. Moreover, during preference alignment, the high overlap between positive and negative instances leads to less informative preference signals, which makes it difficult for the downstream model to learn robust distinctions. Consequently, this configuration achieves the worst performance across datasets. In particular, on LMSYS, the DPO results are even lower than those of SFT.

In contrast, when the thresholds maintain a moderate margin, such as $\tau_{\text{high}} = 0.8$ with $\tau_{\text{low}} = 0.3$ or $\tau_{\text{high}} = 0.9$ with $\tau_{\text{low}} = 0.2$, the model can more reliably identify instances that closely match the ground truth as positives, while assigning instances with clear intent deviation as negatives. These settings induce a clear and well-defined preference boundary and ensure that positive and negative reasoning trajectories remain well separated. As a result, the alignment signal becomes more informative. Empirically, both moderate-margin configurations lead to high-quality preference pairs and achieve the best overall performance.

However, when the threshold gap becomes excessively wide, for example $\tau_{\text{high}} = 1.0$ and $\tau_{\text{low}} = 0.1$, most instances cannot be directly classified and instead must undergo extensive revision or perturbation to generate usable preference pairs. Such strict filtering removes many borderline positive and negative examples that form a smooth spectrum of preference strength. Without exposure to this spectrum, the model loses valuable training diversity, which weakens its generalization ability. As reflected in the results, performance consistently degrades under this extreme configuration.

\subsection{Analysis of Candidate Quantity}
\label{sec:candidates}
This section examines the effect of the number of candidate reasoning trajectories on model performance. By default, ProUtt produces two candidates per perspective, resulting in four candidates in total. To further assess the impact of candidate quantity, we adjust the number of candidates per perspective to 1, 3, and 4, corresponding to overall candidate counts of 2, 6, and 8. As in previous settings, the final performance is determined by selecting the best-scoring candidate among all generated candidates. The results are shown in Appendix Fig. \ref{fig:condidates}.

\par
The LLM-Judge results indicate that model performance does not improve monotonically as the number of candidates increases. Expanding the total candidate count from 2 to 4 yields consistent gains across multiple datasets, suggesting that a moderate increase in candidate diversity provides more informative preference signals. However, further enlarging the candidate pool to 6 or 8 yields saturated or even slightly degraded performance. Examination of the associated reasoning trajectories indicates that additional candidates often add limited semantic variety while introducing more noise, which can undermine the stability of preference learning and bias the model toward inferior patterns. These results indicate that ProUtt primarily benefits from cross-perspective diversity rather than unrestricted expansion within a single perspective, with four candidates providing a favorable trade-off between performance and generation cost. Notably, embedding-based evaluation does not exhibit trends consistent with the LLM-Judge results. Since vector similarity captures surface-level semantic proximity but cannot assess the quality of intent reasoning or trajectory coherence, it is less sensitive to changes in candidate quantity and therefore display more irregular fluctuations.

\section{Conclusion}
This work introduces ProUtt, an LLM-Driven preference data synthesis method for predicting the next user utterance in human–LLM dialogue. ProUtt targets two limitations of existing approaches. User simulators primarily operate at the surface level of utterance generation, while prior data synthesis methods rarely provide explicit reasoning trajectories that distinguish preferred from non-preferred intent paths. Experiments on four datasets, LMSYS, ShareGPT, WildChat and CrossWOZ, evaluated with both LLM and human judgments, show that ProUtt performs better than commercial LLM APIs, simulator-based methods and alternative synthesis approaches. Under LLM-based evaluation, ProUtt achieves relative gains of 4.88, 3.16, 2.46 and 5.12 percent over the strongest baselines on the four datasets. We also provide ablations on core components, backbone LLMs, threshold choices and the number of candidates, and summarize practical settings for different use cases. The ProUtt codebase and synthesized datasets are released to support future research.
\par
Despite its strong performance, our pairwise evaluation results highlight remaining limitations in cross-domain transfer. ProUtt achieves higher win rates on training datasets (LMSYS and CrossWOZ) than on unseen datasets such as ShareGPT and WildChat, indicating room for improvement in generalization. In addition, we model user intent with a tree structure. Graph-based representations can capture richer relational patterns, and exploring them may further improve performance in future work. Moreover, if multiple historical dialogues from the same user are available, incorporating user-specific preference signals into the reasoning process may further enhance personalization and performance.

\bibliographystyle{IEEEtran}
\bibliography{main}



\clearpage
\appendices
\section{Algorithmic Description of ProUtt}
Algorithm~\ref{alg: algorithm} in the appendix provides a concise procedural summary of ProUtt, complementing the conceptual description in Section~\ref{sec:method}.

\begin{algorithm*}[t]
\caption{Preference Data Synthesis Procedure of ProUtt}
\label{alg: algorithm}
\DontPrintSemicolon
\SetKwInOut{Input}{Input}
\SetKwInOut{Output}{Output}

\Input{
Multi-turn dialogue $D^{(N)}=\{d_i\}_{i=1}^{N}$ with $d_i=\langle u_i,a_i\rangle$;\;
$N$: total number of dialogue turns;\;
$Q$: number of candidate next-path proposals per perspective (exploitation/exploration);\;
thresholds $\tau_{\text{low}},\tau_{\text{high}}$.
}
\Output{Preference set $\mathcal{S}$ of tuples $(D^{(K)},T^{(K)},r^+,\hat{U}^+,r^-,\hat{U}^-)$.}

$\mathcal{S}\leftarrow \emptyset$\;
$T^{(N)} \leftarrow \textsc{BuildIntentTree}(D^{(N)})$;\quad
$P^{(N)} \leftarrow \textsc{ExtractNewPaths}(T^{(N)})$\;

\For{$K\leftarrow 1$ \KwTo $N-1$}{
  $D^{(K)},T^{(K)} \leftarrow \textsc{Prefix}(D^{(N)},T^{(N)},K)$;\quad
  $u_{K+1},p_{K+1}\leftarrow \textsc{GTNext}(D^{(N)},P^{(N)},K)$\;

  $(R_{K+1}^{s*},R_{K+1}^{s-}) \leftarrow \textsc{SentenceTypePair}(D^{(K)},u_{K+1})$\;

  $(R_{K+1}^{p},\hat{U}_{K+1}) \leftarrow \textsc{PathReason}(D^{(K)},T^{(K)},R_{K+1}^{s*},Q)$\;
  $J_{\max} \leftarrow \textsc{JudgeMax}(u_{K+1},\hat{U}_{K+1})$\;

  \uIf{$J_{\max}\ge \tau_{\text{high}}$}{
    $R_{K+1}^{p*}\leftarrow R_{K+1}^{p}$;\;
    $p_{K+\varepsilon}\leftarrow \textsc{SampleFuturePath}(P^{(N)},K)$\;
    $R_{K+1}^{p-}\leftarrow \textsc{ReplacePath}(R_{K+1}^{p},p_{K+\varepsilon})$\;
  }
  \uElseIf{$J_{\max}\le \tau_{\text{low}}$}{
    $R_{K+1}^{p-}\leftarrow R_{K+1}^{p}$;\;
    $R_{K+1}^{p*}\leftarrow \textsc{Revise}(R_{K+1}^{p},p_{K+1})$\;
  }
  \Else{
    $R_{K+1}^{p*}\leftarrow \textsc{Revise}(R_{K+1}^{p},p_{K+1})$\;
    $p_{K+\varepsilon}\leftarrow \textsc{SampleFuturePath}(P^{(N)},K)$\;
    $R_{K+1}^{p-}\leftarrow \textsc{ReplacePath}(R_{K+1}^{p},p_{K+\varepsilon})$\;
  }

  $\hat{U}_{K+1}^{+}\leftarrow \textsc{GenerateApprox}(D^{(K)},R_{K+1}^{s*},R_{K+1}^{p*})$\;
  $\hat{U}_{K+1}^{-}\leftarrow \textsc{Generate}(D^{(K)},R_{K+1}^{s-},R_{K+1}^{p-})$\;

  $r^+\leftarrow (R_{K+1}^{s*},R_{K+1}^{p*})$;\quad
  $r^-\leftarrow (R_{K+1}^{s-},R_{K+1}^{p-})$\;
  $\mathcal{S}\leftarrow \mathcal{S}\cup\{(D^{(K)},T^{(K)},r^+,\hat{U}_{K+1}^{+},r^-,\hat{U}_{K+1}^{-})\}$\;
}
\Return $\mathcal{S}$\;
\end{algorithm*}

\section{Experimental Setup}
\label{sec:exp-setup}
\subsection{Implementation Details}
We evaluate the proposed preference data synthesis method ProUtt along two dimensions, language (English and Chinese) and dialogue type (open-domain and task-oriented dialogue). Specifically, we sample dialogue instances from the English open-domain dialogue dataset LMSYS \cite{zheng2023judging} and the Chinese task-oriented dialogue dataset CrossWOZ \cite{zhu2020crosswoz}. These sampled instances are then processed using ProUtt to synthesize preference datasets for next user utterance prediction with explicit reasoning trajectories. The resulting datasets are denoted LMSYS-ProUtt-2K, LMSYS-ProUtt-10K, CrossWOZ-ProUtt-2K and CrossWOZ-ProUtt-5K. The two 2K-scale datasets are used in our experiments, and the larger datasets will be released to the research community. 
\par
Table~\ref{tab:proutt-data-stats} summarizes the statistics of our synthesized preference datasets. 
We report the average text length in characters and in Qwen3-8B tokens, together with the length difference between the preferred and non-preferred responses. 
The mean $\Delta\mathrm{Tokens}$ is close to zero on both datasets ($-2.08$ on CrossWOZ-ProUtt-5K and $3.12$ on LMSYS-ProUtt-10K), suggesting that the preference signal is not dominated by response length. 
Meanwhile, the standard deviation and maximum absolute $\Delta\mathrm{Tokens}$ indicate occasional length mismatches, which are more pronounced on LMSYS-ProUtt-10K (36.78, 407) than on CrossWOZ-ProUtt-5K (22.98, 171).

\begin{table*}[t]
\centering
\caption{Statistics of the synthesized preference data.
Avg.\ Chars and Avg.\ Tokens denote the average text length per field in characters and in Qwen3-8B tokens, respectively.
$\Delta$Tokens denotes the token-length difference between the preferred and non-preferred responses (chosen$-$rejected).
$\mu$ and $\sigma$ indicate the mean and standard deviation of $\Delta$Tokens, and max$|\cdot|$ denotes the maximum absolute value of $\Delta$Tokens.}
\label{tab:proutt-data-stats}
\begin{tabular}{lrrrrrrr}
\toprule
Dataset & \#Samples & Avg.\ Chars & Avg.\ Tokens & $\Delta$Chars & $\Delta$Tokens ($\mu$) & $\Delta$Tokens ($\sigma$) & max$|\Delta\mathrm{Tokens}|$ \\
\midrule
CrossWOZ-ProUtt-5K  & 4{,}980  & 1{,}099.24 & 627.49 & $-6.05$ & $-2.08$ & 22.98 & 171 \\
LMSYS-ProUtt-10K    & 10{,}081 & 3{,}563.63 & 751.11 & 18.69  & 3.12   & 36.78 & 407 \\
\bottomrule
\end{tabular}
\end{table*}

\par
\subsubsection{Training set} In all experiments we use LMSYS-ProUtt-2K and CrossWOZ-ProUtt-2K as training data and fine-tune open-source base LLMs, with Qwen3-8B as the default backbone.
\subsubsection{Test set} Following prior work, we randomly sample 100 test instances from the test splits of LMSYS, ShareGPT \cite{sharegpt_conversations}, WildChat \cite{feuer2025wildchat50m} and CrossWOZ to evaluate downstream models and assess the effectiveness of ProUtt. We do not perform any training on ShareGPT or WildChat, which are only used for out-of-domain generalization. Each configuration is evaluated five times, and we report the average.
\subsubsection{Training setup} We adopt LoRA for parameter-efficient fine-tuning. We first perform 3 epochs of supervised fine-tuning on the chosen responses in the preference datasets. We then run 1 additional epoch of alignment training that uses both chosen and rejected responses. We compare three representative alignment objectives, DPO \cite{rafailov2023dpo}, ORPO \cite{hong-etal-2024-orpo} and SimPO \cite{meng2024simpo}. 
\subsubsection{Hyperparameters} The high-confidence threshold $\tau_{\text{high}}$ is set to 0.8 by default, and the low-confidence threshold $\tau_{\text{low}}$ is set to 0.3. For each perspective we predict two intent-tree paths $(Q = 2)$, so each inference step produces four candidate predictions of the next user utterance. For evaluation, we report, for each metric, the maximum score among the four candidates.

\subsection{Evaluation Metrics}
To comprehensively quantify the performance of ProUtt, we adopt both pointwise and pairwise evaluation protocols.

\subsubsection{Pointwise Evaluation}
First, following prior work~\cite{wang-etal-2025-know, kong2024platolm}, we employ an LLM-as-a-judge protocol \cite{zheng2023judging}. The judge model takes as input the dialogue context, the predicted user utterance and the ground-truth utterance, and assigns a score in the range $[0, 1]$ based on intent consistency and semantic relevance.

Second, following prior work~\cite{wang-etal-2025-know}, we compute the semantic similarity between the predicted and ground-truth utterances using vector-based sentence embeddings. Specifically, we use SimCSE \cite{gao-etal-2021-simcse} for English data and bge-zh-v1.5 \cite{chen2024bge} for Chinese data to measure the semantic closeness of generated utterances.

\subsubsection{Pairwise Evaluation}
For pairwise evaluation, we again follow prior work~\cite{kong2024platolm} and adopt an LLM-as-a-judge setup. Unless otherwise specified, we use Doubao-1.5-Pro as the default judge model. Given the same input, the judge compares the predicted user utterances from two models and decides which one is closer to the ground-truth utterance, yielding win/tie/lose outcomes. To mitigate position bias, we randomly swap the order of the two model predictions.

In addition, we conduct human pairwise evaluation with five recruited annotators. For each evaluation batch of 100 test instances, two annotators independently compare the predictions of the two models and decide which one is closer to the ground-truth user utterance. We treat their judgments as a valid evaluation only if the agreement rate between the two annotators on the batch is at least $0.75$; otherwise the batch is excluded from our analysis. For instances within a valid batch where the two annotators disagree, a third annotator provides the final decision.

\subsection{Baselines}
We compare ProUtt against three classes of baselines: prompt-based LLMs, LLM-based user simulators, and data synthesis methods.

\subsubsection{Prompt-based LLMs}
We evaluate several recent strong LLM baselines, including Qwen3-8B \cite{yang2025qwen3}, Doubao-1.5-Pro \cite{doubao15pro}, Doubao-Seed-1.6 \cite{doubaoseed16}, DeepSeek-V3.2-Exp \cite{deepseekv32}, GLM-4.6 \cite{glm2024chatglm} and Qwen3-Max \cite{yang2025qwen3}. Except for Qwen3-8B, all other models are accessed via their commercial APIs. For each model, we enable its reasoning mode. We adopt a prompting-based setup that directly instructs the model, given the dialogue context, to generate four diverse predictions of the next user utterance in a single inference.

\subsubsection{User Simulators}
We further compare with LLM-based user simulator methods, Socratic \cite{kong2024platolm} and USP \cite{wang-etal-2025-know}. Since Socratic has not been open-sourced, we re-implement its core procedure. USP has been publicly released, and we use its official open-source implementation in our experiments.

\subsubsection{Data Synthesis Methods}
We also consider recent data synthesis methods Baize \cite{xu2023baize}, SPaR \cite{cheng2025spar} and MAGPIE \cite{xu2025magpie}. As these methods are not originally designed for next user utterance prediction, we adapt their prompt templates to this task. All methods are applied to the same raw dialogue data and are configured to generate four candidate predictions of the next user utterance at inference time, to ensure a fair comparison.

\section{Additional Evaluation Results}
This appendix provides supplementary figures and tables that complement the analyses reported in the main text.
\par
\begin{figure*}[!t]
  \centering
  \includegraphics[width=0.8\textwidth]{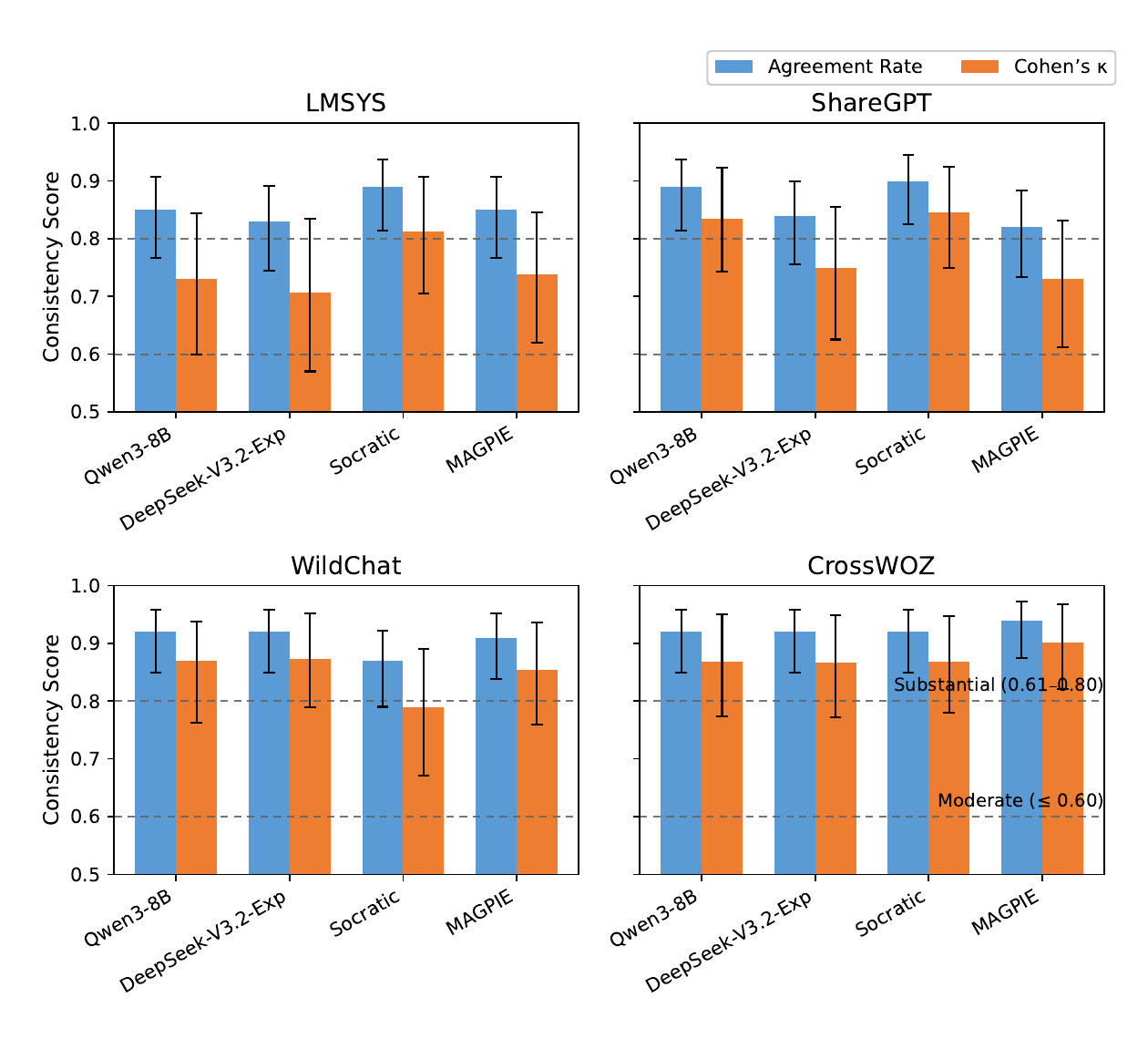}%
  \caption{Human--LLM evaluation consistency across four datasets.
  The bar plots report Agreement Rate and Cohen's $\kappa$ between LLM and human judgments on LMSYS, ShareGPT, WildChat, and CrossWOZ.
  Error bars indicate 95\% confidence intervals for both Agreement Rate and Cohen's $\kappa$.}
  \label{fig:appendix-consistency}
\end{figure*}

Fig.~\ref{fig:appendix-consistency} presents the human--LLM evaluation consistency results discussed in Section~\ref{sec:pairwise_evaluation}, including Agreement Rate, Cohen's $\kappa$, and their 95\% confidence intervals across the four datasets.
\par
\begin{figure*}[!t]
  \centering
  \includegraphics[width=0.8\textwidth]{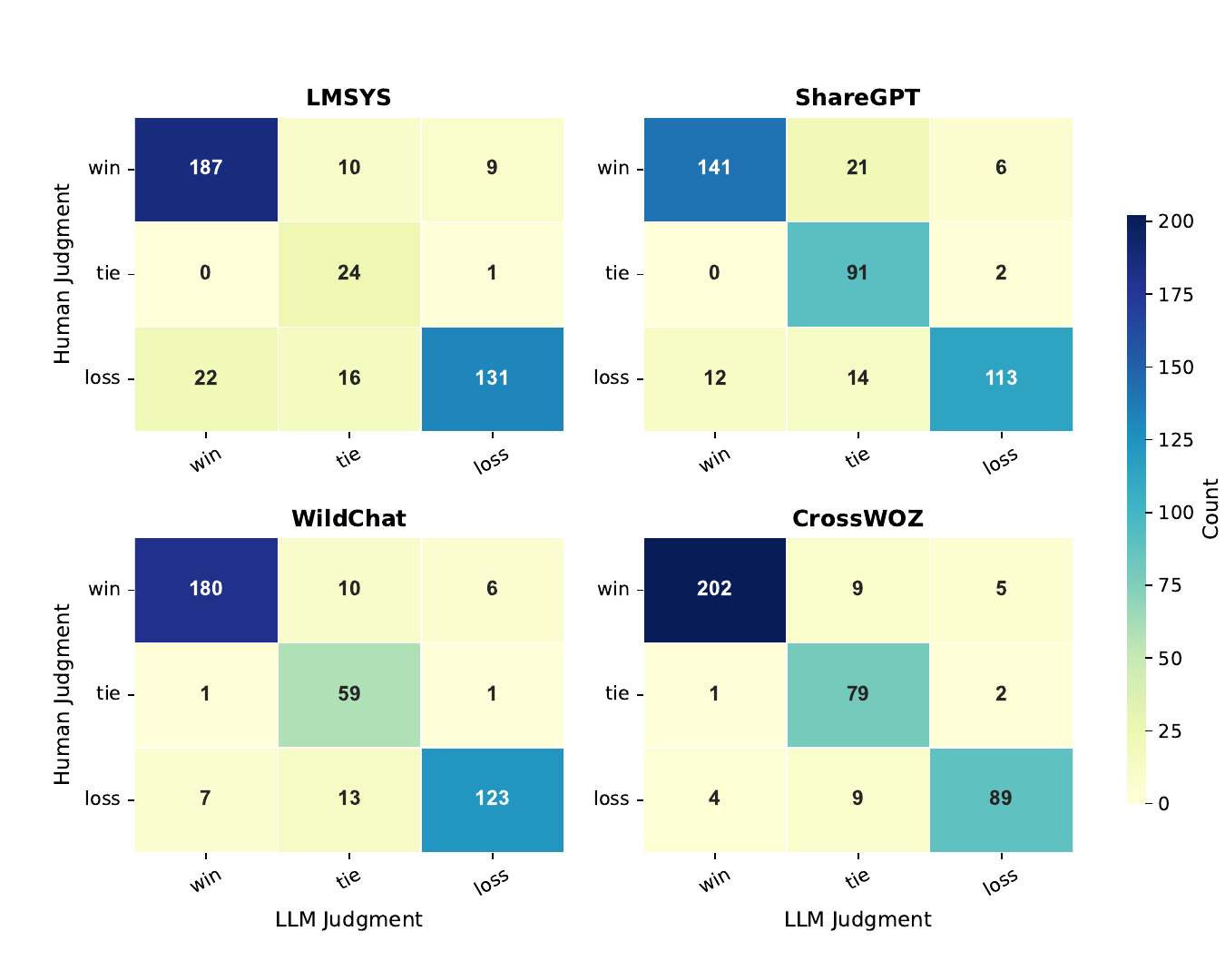}%
  \caption{Confusion matrices comparing human and LLM pairwise judgments across four datasets, illustrating their agreement over win, tie, and loss outcomes.
  Darker diagonal cells indicate higher agreement and thus stronger consistency between LLM-based and human evaluations.}
  \label{fig:appendix-confusion}
\end{figure*}

Fig.~\ref{fig:appendix-confusion} presents the confusion matrices referenced in Section~\ref{sec:pairwise_evaluation}, comparing human and LLM pairwise judgments across four datasets.
\par
Table~\ref{tab:ablation} presents the ablation study results of ProUtt. Specifically, “w/o Intent Tree” removes the explicit intent-tree construction module, “w/o Sentence Type Reasoning” eliminates the sentence type reasoning step, and “w/o Future-Path Negatives” replaces the future-path-based negative construction strategy with LLM-generated negative paths.
\par
Table~\ref{tab:llm-backbone} presents the detailed results for the analysis in Section~\ref{sec:llm-backbone}, evaluating how different LLM backbones affect the quality of ProUtt-generated data.
\par
Table~\ref{tab:backbone-sensitivity} presents the results discussed in Section~\ref{sec:backbone}, showing how backbone LLMs of different scales and model families respond to ProUtt-generated synthetic data across four evaluation benchmarks.
\par
Table~\ref{tab:instruction-structure} presents the results discussed in Section~\ref{sec:instruction}, comparing model performance under structured versus minimal instruction formats.
\par
Table~\ref{tab:threshold-sensitivity} presents the results discussed in Section~\ref{sec:threshold}, analyzing how different configurations of the high- and low-confidence thresholds affect ProUtt’s performance under both SFT and DPO training settings.
\par
\begin{figure*}
    \centering
    \includegraphics[width=1\linewidth]{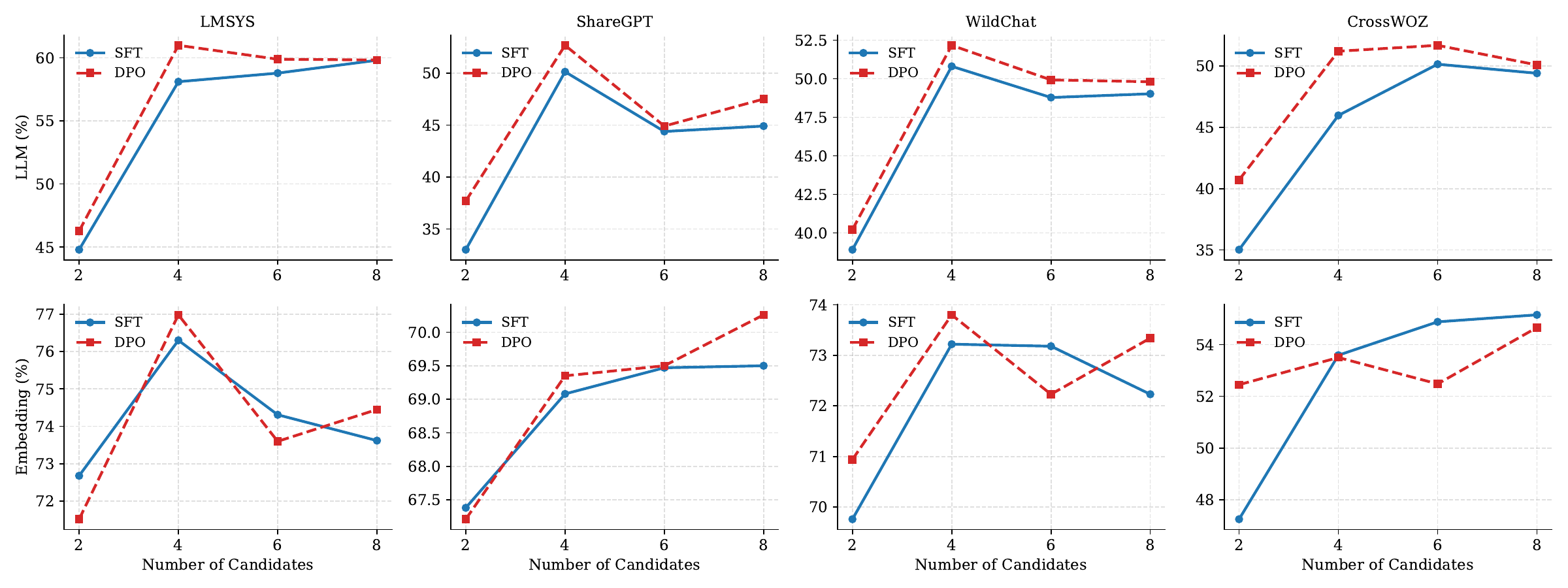}
    \caption{Effect of the number of candidates generated by ProUtt on model performance across four datasets. Performance is reported using the best-scoring candidate under each setting.}
    \label{fig:condidates}
\end{figure*}

Fig.~\ref{fig:condidates} presents the results discussed in Section~\ref{sec:candidates}, illustrating how different candidate quantities influence model performance across four datasets.

\begin{table*}[t]
\centering
\small
\caption{Ablation study results of ProUtt.
“w/o Intent Tree” removes the explicit intent-tree construction module. 
“w/o Sentence Type Reasoning” removes the sentence type reasoning step. 
“w/o Future-Path Negatives” replaces the negative construction strategy based on future intent-tree paths with LLM-generated negative paths.}
\label{tab:ablation}
\resizebox{\linewidth}{!}{
\begin{tabular}{lcccccccc}
\toprule
\multirow{2}{*}{Module} & \multicolumn{2}{c}{LMSYS} & \multicolumn{2}{c}{ShareGPT} & \multicolumn{2}{c}{WildChat} & \multicolumn{2}{c}{CrossWOZ} \\
\cmidrule(lr){2-3}
\cmidrule(lr){4-5}
\cmidrule(lr){6-7}
\cmidrule(lr){8-9}
 & LLM-Judge & Embed-Sim & LLM-Judge & Embed-Sim & LLM-Judge & Embed-Sim & LLM-Judge & Embed-Sim \\
\midrule
\multicolumn{9}{l}{\textbf{SFT}} \\
ProUtt & 58.10 & 76.30 & 50.12 & 69.08 & 50.80 & 73.22 & 45.98 & 53.59 \\
w/o Intent Tree & 50.80 & 74.34 & 40.14 & 67.71 & 40.94 & 71.03 & 45.80 & 53.70 \\
w/o Sentence Type Reasoning & 53.78 & 74.56 & 45.72 & 68.39 & 47.74 & 72.64 & 44.04 & 52.69 \\
w/o Future-Path Negatives & 55.22 & 74.97 & 44.82 & 69.66 & 45.76 & 70.42 & 42.52 & 52.83 \\
\midrule
\multicolumn{9}{l}{\textbf{DPO}} \\
ProUtt & 60.98 & 76.98 & 52.66 & 69.35 & 52.16 & 73.80 & 51.22 & 53.50 \\
w/o Intent Tree & 52.30 & 74.55 & 41.76 & 67.85 & 42.50 & 71.19 & 47.74 & 52.83 \\
w/o Sentence Type Reasoning & 53.80 & 74.18 & 45.82 & 69.22 & 48.90 & 72.87 & 44.64 & 51.22 \\
w/o Future-Path Negatives & 56.46 & 73.33 & 44.94 & 67.91 & 46.32 & 71.97 & 43.56 & 52.39 \\
\bottomrule
\end{tabular}
}
\end{table*}

\begin{table*}[t]
\centering
\caption{Performance of ProUtt with different backbone LLMs. Metrics that exceed the default Doubao-1.5-Pro configuration are highlighted in bold in the main text.}
\label{tab:llm-backbone}
\small
\begin{tabular}{lcccccccc}
\toprule
\multirow{2}{*}{LLM} & \multicolumn{2}{c}{LMSYS} & \multicolumn{2}{c}{ShareGPT} & 
\multicolumn{2}{c}{WildChat} & \multicolumn{2}{c}{CrossWOZ} \\
\cmidrule(lr){2-3} \cmidrule(lr){4-5} \cmidrule(lr){6-7} \cmidrule(lr){8-9}
& LLM-Judge & Embed-Sim & LLM-Judge & Embed-Sim & LLM-Judge & Embed-Sim & LLM-Judge & Embed-Sim\\
\midrule
\multicolumn{9}{l}{\textbf{SFT}} \\
Doubao-1.5-Pro & 58.10 & 76.30 & 50.12 & 69.08 & 50.80 & 73.22 & 45.98 & 53.59 \\
Qwen3-Plus       & 55.42 & 73.82 & 46.28 & \textbf{69.42} & 47.44 & 72.41 & \textbf{46.64} & \textbf{55.33} \\
DeepSeek-V3.2-Exp   & \textbf{58.28} & 73.83 & 48.50 & \textbf{69.33} & 47.32 & 72.45 & 43.40 & 52.57 \\
GLM-4.5-Air         & 56.78 & 74.38 & 47.52 & \textbf{69.37} & 47.50 & 72.44 & 43.86 & 53.06 \\
Range (Max$-$Min)     & 2.86  & 2.48  & 3.84  & 0.34  & 3.48  & 0.81  & 3.24  & 2.76 \\
\midrule
\multicolumn{9}{l}{\textbf{DPO}} \\
Doubao-1.5-Pro & 60.98 & 76.98 & 52.66 & 69.35 & 52.16 & 73.80 & 51.22 & 53.50 \\
Qwen3-Plus       & 56.78 & 73.92 & 49.12 & 68.62 & 48.44 & 72.24 & 48.30 & \textbf{55.53} \\
DeepSeek-V3.2-Exp       & 60.10 & 73.03 & 49.22 & 69.29 & 50.90 & \textbf{74.02} & 50.68 & 53.04 \\
GLM-4.5-Air         & 59.22 & 73.02 & 49.50 & 68.73 & 49.28 & \textbf{73.87} & 50.02 & \textbf{53.64} \\
Range (Max$-$Min)        & 4.20  & 3.96  & 3.54  & 0.73  & 3.72  & 1.78  & 2.92  & 2.49 \\
\bottomrule
\end{tabular}
\end{table*}

\begin{table*}[t]
\centering
\small
\caption{Performance of different backbone LLMs fine-tuned with ProUtt-generated synthetic data.}
\label{tab:backbone-sensitivity}
\begin{tabular}{lcccccccc}
\toprule
\multirow{2}{*}{Model} & 
\multicolumn{2}{c}{LMSYS} & 
\multicolumn{2}{c}{ShareGPT} & 
\multicolumn{2}{c}{WildChat} & 
\multicolumn{2}{c}{CrossWOZ} \\
\cmidrule(lr){2-3} \cmidrule(lr){4-5} \cmidrule(lr){6-7} \cmidrule(lr){8-9}
& LLM-Judge & Embed-Sim & LLM-Judge & Embed-Sim & LLM-Judge & Embed-Sim & LLM-Judge & Embed-Sim\\
\midrule
\multicolumn{9}{l}{\textbf{SFT}} \\
Qwen3-0.6B   & 52.62 & 70.70 & 36.46 & 66.00 & 39.36 & 69.70 & 43.90 & 53.06 \\
Qwen3-1.7B   & 53.34 & 73.18 & 39.70 & 68.63 & 42.92 & 70.86 & 44.22 & 51.85 \\
Qwen3-4B     & 54.50 & 74.45 & 39.82 & 67.91 & 44.60 & 72.15 & 45.76 & 52.46 \\
Qwen3-8B     & 58.10 & 76.30 & 50.12 & 69.08 & 50.80 & 73.22 & 45.98 & 53.59 \\
LLaMA3.1-8B  & 55.60 & 73.16 & 44.58 & 69.03 & 43.62 & 70.57 & 44.06 & 52.62 \\
\midrule
\multicolumn{9}{l}{\textbf{DPO}} \\
Qwen3-0.6B   & 53.10 & 72.57 & 37.02 & 67.34 & 41.92 & 69.18 & 45.60 & 52.19 \\
Qwen3-1.7B   & 54.62 & 73.29 & 40.92 & 68.29 & 43.96 & 70.74 & 47.38 & 52.33 \\
Qwen3-4B     & 55.12 & 74.82 & 42.04 & 68.28 & 45.50 & 72.53 & 48.56 & 52.67 \\
Qwen3-8B     & 60.98 & 76.98 & 52.66 & 69.35 & 52.16 & 73.80 & 51.22 & 53.50 \\
LLaMA3.1-8B  & 61.12 & 74.58 & 45.50 & 68.47 & 46.10 & 70.71 & 51.20 & 55.19 \\
\bottomrule
\end{tabular}
\end{table*}

\begin{table*}[t]
\centering
\small
\caption{Performance comparison between structured and minimal instruction formats.}
\label{tab:instruction-structure}
\begin{tabular}{lcccccccc}
\toprule
& \multicolumn{2}{c}{LMSYS} & \multicolumn{2}{c}{ShareGPT} & \multicolumn{2}{c}{WildChat} & \multicolumn{2}{c}{CrossWOZ} \\
\cmidrule(lr){2-3} \cmidrule(lr){4-5} \cmidrule(lr){6-7} \cmidrule(lr){8-9}
& LLM-Judge & Embed-Sim & LLM-Judge & Embed-Sim & LLM-Judge & Embed-Sim & LLM-Judge & Embed-Sim\\
\midrule
\textbf{SFT} \\
Structured & 58.10 & 76.30 & 50.12 & 69.08 & 50.80 & 73.22 & 45.98 & 53.59 \\
Minimal    & 55.22 & 73.66 & 45.78 & 68.67 & 48.92 & 72.45 & 44.82 & 52.39 \\
\midrule
\textbf{DPO} \\
Structured & 60.98 & 76.98 & 52.66 & 69.35 & 52.16 & 73.80 & 51.22 & 53.50 \\
Minimal    & 57.92 & 73.74 & 48.32 & 69.66 & 49.34 & 72.68 & 50.55 & 52.78 \\
\bottomrule
\end{tabular}
\end{table*}

\begin{table*}[t]
\centering
\small
\caption{Threshold sensitivity analysis for ProUtt under different choices of the high-confidence threshold $\tau_{\text{high}}$ and the low-confidence threshold $\tau_{\text{low}}$. Results are reported under both SFT and DPO training settings.}
\label{tab:threshold-sensitivity}
\renewcommand{\arraystretch}{1.1}
\begin{tabular}{cc|cccccccc}
\toprule
\multicolumn{2}{c|}{Thresholds} &
\multicolumn{2}{c}{LMSYS} &
\multicolumn{2}{c}{ShareGPT} &
\multicolumn{2}{c}{WildChat} &
\multicolumn{2}{c}{CrossWOZ} \\
\cmidrule(lr){1-2} \cmidrule(lr){3-4} \cmidrule(lr){5-6}
\cmidrule(lr){7-8} \cmidrule(lr){9-10}
High & Low & LLM-Judge & Embed-Sim & LLM-Judge & Embed-Sim & LLM-Judge & Embed-Sim & LLM-Judge & Embed-Sim\\
\midrule
\multicolumn{10}{l}{\textbf{SFT}} \\
0.7 & 0.4 & 55.38 & 73.51 & 44.02 & 69.06 & 45.04 & 71.65 & 41.50 & 50.68 \\
0.8 & 0.3 & 58.10 & 76.30 & 50.12 & 69.08 & 50.80 & 73.22 & 45.98 & 53.59 \\
0.9 & 0.2 & 56.62 & 76.44 & 48.64 & 69.91 & 48.42 & 72.60 & 44.80 & 51.69 \\
1.0 & 0.1 & 57.22 & 73.71 & 46.40 & 68.62 & 45.04 & 71.31 & 43.82 & 52.59 \\
\midrule
\multicolumn{10}{l}{\textbf{DPO}} \\
0.7 & 0.4 & 53.02 & 73.36 & 47.44 & 68.76 & 45.10 & 70.58 & 43.58 & 51.57 \\
0.8 & 0.3 & 60.98 & 76.98 & 52.66 & 69.35 & 52.16 & 73.80 & 51.22 & 53.50 \\
0.9 & 0.2 & 60.07 & 76.83 & 49.02 & 69.35 & 50.90 & 73.59 & 51.28 & 54.46 \\
1.0 & 0.1 & 58.80 & 74.00 & 47.40 & 68.91 & 47.14 & 72.22 & 47.78 & 53.99 \\
\bottomrule
\end{tabular}
\end{table*}

\section{Preference Data Examples}
Fig. ~\ref{fig:appendix-pref-example-chosen} and Fig. ~\ref{fig:appendix-pref-example-rejected} show a representative preference data instance constructed by ProUtt, including the dialogue context, the ground-truth next user utterance, and the chosen and rejected reasoning trajectories and responses.

\begin{figure*}[t]
    \centering
    \small
    \setlength{\tabcolsep}{4pt}
    \renewcommand{\arraystretch}{1.2}
    \begin{tabularx}{0.95\textwidth}{@{}>{\raggedright\arraybackslash}X@{}}
        \toprule
        \textbf{Context (Dialogue History)} \\
        \textbf{User:} \textit{Convert this short story to first person point of view. Story: Once upon a time, in a big forest, there lived a rhinoceros named NAME\_1. NAME\_1 loved to climb. \ldots{} And NAME\_1 learned that with a little help from a friend, she could climb anything.} \\
        \textbf{LLM:} \textit{I am NAME\_1, the rhinoceros. I am living in a big, beautiful forest. One day, I found an icy hill that I had never seen before. I tried to climb it, but it was very slippery. I tried again and again, but I kept falling down. I was so sad. \ldots{} And NAME\_1 learned that with a little help from a friend, she could climb anything.} \\
        \midrule
        \textbf{Label (Ground-Truth Next User Utterance)} \\
        \textit{Convert this to second person point of view.} \\[0.5ex]
        \midrule
        \textbf{Chosen CoT (Reasoning)} \\
        \textit{First, I need to extract user intent from the historical conversation between the user and the assistant and represent it using a tree structure.} \\[0.3ex]
[User Intent Tree]:\\
StoryConversion \{\\
\quad OriginalPOV: Third person,\\
\quad TargetPOV: First person,\\
\quad Story: Rhinoceros NAME\_1 tries to climb icy hill, fails, gets help from bird NAME\_2, climbs it and they become friends.\\
\}
\\[0.5ex]
        \textit{Next, I need to analyze the sentence type of the user's next input.} \\[0.3ex]
[Sentence Type Analysis]: The user may issue an instruction based on the converted story. They could ask the assistant to make further modifications, such as changing the tone of the story, adding more details, or converting it to a different style. They might also ask for another related task, such as producing a summary of the first-person version. Therefore, the sentence type for the user's next input is most likely Instruction.\\[0.5ex]
        \textit{Finally, I need to infer potential inputs that may interest the user based on the intent tree, considering both the mining and exploration views.} \\[0.3ex]
        [Mining View Analysis]: In the mining view, we focus on identifying new attributes or modifying existing attribute values under the current StoryConversion topic. The user has already converted a story from third person to first person. One possible new attribute is the output style, such as converting the story to a more formal style. The corresponding intent tree mining path would be: StoryConversion $\rightarrow$ OutputStyle - Formal. Another option is to change the target point of view to second person. The intent tree mining path for this would be: StoryConversion $\rightarrow$ TargetPOV - Second person. These predictions follow from the user's interest in story conversion and the potential for further adjustments within the same topic. \\[0.5ex]
        [Exploration View Analysis]: For the exploration view, we look for new topics that are closely related to the existing StoryConversion topic. One such new topic is story summarization. After converting the story, the user might want to summarize it to obtain a concise version, which leads to the exploration path StoryConversion $\rightarrow$ StorySummarization - SummaryLength. Another new topic is story translation. Since the user is working with a story, they might want to translate it into another language, yielding the exploration path StoryConversion $\rightarrow$ StoryTranslation - TargetLanguage. These predictions expand on the user's engagement with the story beyond conversion alone. 
        \\[0.5ex]
        \midrule
        \textbf{Chosen Response} \\
        \textit{The user's next input is most likely one of the following:} \\
        1.\ Convert the story to a formal style. \\
        2.\ Convert the story to second person point of view. \\
        3.\ Summarize the story. \\
        4.\ Translate the story into another language. \\[1ex]
        \bottomrule
    \end{tabularx}
    \caption{Preference data example (Part 1): dialogue context, ground-truth next user utterance, and the chosen reasoning trajectory and response.}
    \label{fig:appendix-pref-example-chosen}
\end{figure*}

\begin{figure*}[t]
    \centering
    \small
    \setlength{\tabcolsep}{4pt}
    \renewcommand{\arraystretch}{1.2}
    \begin{tabularx}{0.95\textwidth}{@{}>{\raggedright\arraybackslash}X@{}}
        \toprule
        \textbf{Rejected CoT (Reasoning)} \\
        \textit{First, I need to extract user intent from the historical conversation between the user and the assistant and represent it using a tree structure.} \\[0.5ex]
[User Intent Tree]:\\
StoryConversion \{\\
\quad OriginalPOV: Third person,\\
\quad TargetPOV: First person,\\
\quad Story: Rhinoceros NAME\_1 tries to climb icy hill, fails, gets help from bird NAME\_2, climbs it and they become friends.\\
\}
\\[0.5ex]

        \textit{Then, I need to analyze the sentence type of the user's next input.} \\[0.5ex]
        [Sentence Type Analysis]: The user may provide a statement about their thoughts on the converted story, such as expressing their opinion on the quality of the conversion, how much they like the first-person version, or giving general feedback about the story's content from the new perspective. Since the assistant has just completed the conversion task, the user might want to share their feelings or observations in a non-interrogative and non-directive way. Therefore, the sentence category for the user's next input is most likely Statement. \\[0.8ex]

        \textit{Finally, I need to infer potential inputs that may interest the user based on the intent tree, considering both mining and exploration views.} \\[0.5ex]
        [Mining View Analysis]: In the mining view, we focus on finding new attributes or changing existing attribute values under the current StoryConversion topic. The user has already converted a story from third person to first person. One possible new attribute is the author name of the story, which gives the mining path StoryConversion $\rightarrow$ AuthorName. Another option is to change the original point of view to No person, which corresponds to the mining path StoryConversion $\rightarrow$ OriginalPOV: No person. These predictions are based on the user's interest in story conversion and the potential for further adjustments within the same topic. \\[0.8ex]

        [Exploration View Analysis]: For the exploration view, we look for new topics that are closely related to the existing StoryConversion topic. One such new topic is story summarization. After converting the story, the user might want to summarize it to obtain a concise version, which leads to the exploration path StoryConversion $\rightarrow$ StorySummarization - SummaryLength. Another new topic is story translation. Since the user is working with a story, they might want to translate it into another language, yielding the exploration path StoryConversion $\rightarrow$ StoryTranslation - TargetLanguage. These predictions expand on the user's engagement with the story beyond conversion alone. \\[0.8ex]

        \midrule
        \textbf{Rejected Response} \\
        \textit{The user's next input is most likely one of the following:} \\
        1.\ I want to know the author of this story. \\
        2.\ I want to convert the story with the original point of view as no person to first person. \\
        3.\ Summarize the story. \\
        4.\ Translate the story into another language. \\[1ex]
        \bottomrule
    \end{tabularx}
    \caption{Preference data example (Part 2): rejected reasoning trajectory and response corresponding to Fig.~\ref{fig:appendix-pref-example-chosen}.}
    \label{fig:appendix-pref-example-rejected}
\end{figure*}

\section{Case Study}

Fig.~\ref{fig:case-study-rust} presents an illustrative case that compares next user utterance predictions from different methods under the same dialogue context.

\begin{figure*}[t]
\centering
\small
\setlength{\tabcolsep}{4pt}
\renewcommand{\arraystretch}{1.2}
\begin{tabularx}{0.95\textwidth}{@{}>{\raggedright\arraybackslash}X@{}}
\toprule

\textbf{Context (Dialogue History)} \\
\textbf{User:} \codestr{Hi, can you explain the fallowing rust code in details! \n
```Rust\n#[derive(Debug, Clone)]\npub struct Person {\n    name: NAME_1,\n    age: i32,\n}\n\nimpl Person {\n    pub fn new(name: NAME_1, age: i32) -> Self {\n        Person { name, age }\n    }\n}\n\nfn main() {\n    println!(\"Hello, world!\");\n}\n```} \\
\textbf{Assistant:} \codestr{Sure! This Rust code defines a struct called `` Person `` that represents a person with a name and an age. The `` Person `` struct is defined using the `` derive `` macro, which allows you to automatically derive default implementations of common traits such as `` Debug `` and `` Clone `` based on the properties of the struct.\n\nThe `` Person `` struct has two fields: `` name `` and `` age ``. The `` name `` field is a NAME_1, and the `` age `` field is an integer. The `` new `` method is used to create a new `` Person `` instance with a specified name and age.\n\nThe `` main `` function simply prints a message to the console.\n\nI hope that helps! Let me know if you have any specific questions about this code or if there's anything else you'd like me to explain.} \\

\\[-0.3em]
\midrule
\textbf{Label (Ground-Truth Next User Utterance)} \\
Why Rust use \texttt{\#[derive(Debug, Clone)]}? \\
\\[-0.3em]

\midrule

\textbf{Predicted Next User Utterances (Top-4)} \\

\textit{ProUtt:}
\begin{enumerate}
  \item What is the purpose of the \texttt{derive} macro in this Rust code?
  \item What does the \texttt{Debug} trait do in this code?
  \item How can I optimize this Rust code?
  \item How can I test the \texttt{Person} struct and the \texttt{new} method?
\end{enumerate}

\textit{Qwen3-8B:}
\begin{enumerate}
  \item What is the \texttt{NAME\_1} type in the code?
  \item Can you explain the purpose of deriving the \texttt{Debug} and \texttt{Clone} traits?
  \item How can I modify this code to add a method that calculates the person's age?
  \item Can you show an example of how to use the \texttt{Person} struct in \texttt{main}?
\end{enumerate}

\textit{DeepSeek-V3.2-Exp:}
\begin{enumerate}
  \item What is \texttt{NAME\_1} in the code?
  \item Why do we need the \texttt{Debug} and \texttt{Clone} traits for this struct?
  \item Could you show an example of how to create and use a \texttt{Person} instance?
  \item How would I add methods to update the age field?
\end{enumerate}

\textit{MAGPIE:}
\begin{enumerate}
  \item What is the \texttt{NAME\_1} type in the code?
  \item Can you explain the \texttt{derive} macro in more detail?
  \item What is the purpose of the \texttt{new} method?
  \item What does the \texttt{main} function do in this code?
\end{enumerate}

\\[0.5em]
\bottomrule
\end{tabularx}
\caption{Case study comparing next-user utterance predictions across models for the same dialogue context.}
\label{fig:case-study-rust}
\end{figure*}

\end{document}